\documentclass[sensors,article,accept,moreauthors,pdftex]{Definitions/mdpi} 
\usepackage[]{algorithm2e}
\usepackage{amsmath}
\usepackage{tabularx}
\usepackage{multicol}
\usepackage{comment}
\usepackage{array}
\usepackage{mathtools}
\usepackage{upgreek} 
 \usepackage{textcomp}

\newcommand{\disclaimer}[1]{%
\vspace{6pt}\noindent{\fontsize{9}{9}\selectfont\textbf{Disclaimer:} {#1}\par}}
\firstpage{1} 
\pubvolume{xx}
\issuenum{1}
\articlenumber{5}
\pubyear{2019}
\copyrightyear{2019}
\history{Received: date; Accepted: date; Published: date}
\updates{yes} % If there is an update available, un-comment this line

\Title{An FPGA-Based Neuro-Fuzzy Sensor for Personalized Driving Assistance}

\Author{{Óscar Mata-Carballeira} %Please carefully check the accuracy of names and affiliations.
*\orcidA{}, {Jon Gutiérrez-Zaballa}, Inés del Campo and Victoria Martínez}

\AuthorNames{Óscar Mata-Carballeira, Jon Gutiérrez-Zaballa, Inés del Campo and Victoria Martínez}

\address[1]{Department of Electricity and Electronics, Faculty of Science and Technology, University of the Basque Country UPV/EHU, 48940 Leioa, Spain; jongutierrez@bizkaia.eu (J.G.-Z.); ines.delcampo@ehu.eus (I.d.C.); victoria.martinez@ehu.eus (V.M.)}

\corres{\hangafter=1 \hangindent=1.05em \hspace{-0.82em} {Correspondence: oscar.mata@ehu.eus; Tel.: +34-94-601-3238}}

\abstract{Advanced driving-assistance systems (ADAS) are intended to automatize driver tasks, as well as improve  driving and vehicle safety. This work proposes an intelligent neuro-fuzzy sensor for driving style (DS) recognition, suitable for ADAS enhancement. The development of the driving style intelligent sensor uses naturalistic driving data from the SHRP2 study, which includes data from a CAN bus, inertial measurement unit, and front radar. The system has been successfully implemented using a field-programmable gate array (FPGA) device of the Xilinx Zynq programmable system-on-chip (PSoC). It can mimic the typical timing parameters of a group of drivers as well as tune these typical parameters to model individual DSs. The neuro-fuzzy intelligent sensor provides high-speed  real-time active ADAS implementation and is able to personalize its behavior into safe margins without driver intervention. In particular, the personalization procedure of the time headway (THW) parameter for an ACC in steady car following was developed, achieving a performance of 0.53 microseconds. This performance fulfilled the requirements of cutting-edge active ADAS specifications.%The characterization of the several driving behaviors has been made offline by clustering characteristic parameters. This clustering has been performed by applying the k-means algorithm on previously identified steady car-following segments from the SHRP2 naturalistic study. 
}

\keyword{advanced driving assistance systems (ADAS); safety and comfort; driving~style; unsupervised clustering; k-means; adaptive neuro-fuzzy inference system (ANFIS); field-programmable gate array (FPGA); programmable system-on-chip (PSoC)}

\begin{document}
%%%%%%%%%%%%%%%%%%%%%%%%%%%%%%%%%%%%%%%%%%

%%%%%%%%%%%%%%%%%%%%%%%%%%%%%%%%%%%%%%%%%%

\section{Introduction}
Currently, manufactured cars rely on a myriad of sensors to measure many  internal and external variables that could influence the handling behavior of the automobile, as well as additional parameters, such as visibility and occupant comfort. Depending on their sophistication level, sensors can be classified ranging from simple sensors that directly measure single physical parameters (e.g., ambient light sensors and temperature sensors) to complex intelligent sensors, which determine parameters of the surrounding environment through wide spectrum signals (e.g., radio frequency/radar and light/video); besides measuring, they perform data processing and are enabled to carry out actuations. Whereas intelligent sensors make use of data of a different nature underneath, in which complex and nonlinear behaviors are codified; data-mining techniques used jointly with machine learning (ML) algorithms have shown adequate performance for modeling this hidden information. As~intelligent sensors often rely on complex sensors and sensor fusion techniques, the data processing power they need can  only be provided by high-performance computational platforms such as microprocessors, graphics-processing units (GPUs), or field-programmable gate arrays (FPGAs). In~particular, FPGA-based implementations stand out due to the extremely high operational frequencies and low power consumption they can achieve, even for complex, multilayered algorithms \cite{fernandes2}. In the context of the automotive field, intelligent sensors are key components of current assistance systems. 

A long time has passed since automobiles were simple mechanical systems. With the development of the first driving-assistance systems (DAS) in the 1980s, cars introduced electronics in daily driving, turning them into complex mechatronic systems \cite{DAS}. Thus, they have been gradually fitted with several simple electronic sensors, such as angular speed sensors, torque sensors, accelerometers, and gyroscopes, which together with mechatronic drives have derived safety systems such as the anti-lock braking system (ABS) \cite{leiber1980antiblockiersystem}, traction-control system (TCS) \cite{bleckmann1986traction}, and electronic stability program (ESP)~\cite{farmer2004effect}. These systems, before being marketed  as optional equipment in top tier models,  have drastically improved automotive safety. This improvement led to the obligation of fitting these systems in every new car in the United States of America (USA) since 2011 \cite{DASreguUSA} and in the European Union (EU) since~2014~\cite{DASregu}. 

In the field of driver comfort, simple speed sensors play an important role in conventional feedback systems, such as cruise control (CC) solutions that relieve drivers from the mental load of keeping  speed in a steady state during long rides in highways and motorways. However, despite CC systems contributing to reducing drivers' fatigue during long trips, they may bring about dangerous situations, especially rear-end collisions \cite{smiley2000behavioral}. These situations occur because conventional CC systems are not aware of the distance from the preceding vehicle, that is, they only take into account the internal variables of the car. To solve these problems, external world data-based systems, such as adaptive cruise control (ACC), arose as an alternative to provide enhanced safety and comfort functionalities. ACC relies on real-time measurements of the distance to the preceding vehicle to keep the time gap and avoid rear-end collisions, thus increasing road safety \cite{schleicher2011influence}. Systems like ACC are based on complex intelligent sensors that measure external parameters through high-bandwidth signals such as radar, LIDAR, or video. These kinds of enhanced systems are known as advanced driving-assistance systems (ADAS) \cite{DAS}.

 ADAS rely on a continuous stream of data from multiple sensors that measure internal and external variables to provide advanced functionalities \cite{feilhauer2016current}. They can be classified into two categories taking into account their actuation level: passive and active ADAS. The former provides advice or information to the driver. Examples of passive ADAS are blind-spot sensors that use ultrasound sensors to detect obstacles  in the blind spot of the rear-view mirrors. Collision-avoidance systems use radar, and seldom LIDAR or video, to detect potential front collisions, warning the driver (forward-collision warning (FCW)) \cite{ho2007multisensory}. Lane-departure warning systems (LDW) \cite{mahajan2015lane} detect  lane marks by video signals and inform the driver about lane-departure events. Finally, traffic sign identification/recognition systems (TSI/TSR) are able to detect speed limits displaying warnings in the dashboard of the vehicle. The latter, active ADAS, can perform actions on the car, and includes systems such as autonomous emergency braking (AEB) \cite{ho2007multisensory}, which can automatically stop the car. Lane-keeping assistance (LKA)~\cite{mahajan2015lane} is a step forward from LDW systems; it can correct the trajectory of the vehicle by autosteering manoeuvres to avoid unintended lane changes. Other examples of active ADAS are the above introduced ACC and automatic speed assistants (ASA), which, based on TSI/TSR and jointly acting with a global positioning system (GPS) signal, automatically apply the speed limit of the road to the car \cite{wali2015comparative}. 
 
 The aforementioned systems have been demonstrated to improve safety in cars since they relieve drivers from some of the most safety-critical tasks \cite{caber2019designing}. The conjunction of all of these intelligent sensors, together with high-speed wireless communications, have allowed car-makers for the first time to develop intelligent vehicles with automated driving systems \cite{murray2017inside}. 
The acceptance of ADAS by drivers mainly depends on the engineering factors of the system, predefined by  technicians and implied by  system functional specifications \cite{fleming2018adaptive}. Thus, for the longitudinal control of vehicles, the time gap with the precedent vehicle, known as the time headway (THW), is monitored by means of radar sensors, considering the longitudinal area of safe travel in relation to the predefined stop distance parameter limits set by the manufacturer. However, despite these systems being very easy to use and contribute to improving road safety,  acceptance by the motorists has been limited \cite{beggiato2015learning}. This lack of acceptance is caused by drivers perceiving the car as not as natural and stable as expected, despite it being programmed to be as conservative and safe as possible. Furthermore, FCW systems' conservative THW safety parameters may make some drivers feel distrust on the systems and annoyance, as they return an excessive number of warnings, making motorists prone to dismiss this advice or even to disengage the system \cite{panou2018intelligent}. This lack of use causes their effect on global sinistrality rates to not be as good as previously foreseen, despite ACC systems having been gradually installed in new production cars and demonstrated to contribute in reducing accidents' severity, even eliminating them \cite{piccinini2014driver}.

%\textbf{Personalization in Advanced Driver Assistance Systems and Autonomous Vehicles: A Review}
\subsection*{Personalization Approaches in ADAS}

Several car-makers have introduced a slight level of ADAS customization  by allowing users to manually select the THW parameter from a knob or a lever in the steering wheel, always within the pre-engineered parameters. However, despite a group of drivers appreciating the freedom that manually adjustable parameters provide them in automatic longitudinal control modes; this personalization process, including the operation of the system itself, can be complicated for other drivers not so familiarized with control systems in automobiles. Therefore, it seems reasonable to introduce personalization strategies that need no driver intervention to make the adoption of these systems easier for that sector of the automotive community \cite{hasenjager2017personalization}. ADAS personalization embeds characteristics of  motorists' driving style into the system. The driving style (DS) is the manner the driver operates a vehicle in terms of steering, acceleration and braking, and how this driver relates to the other ones in terms of predictability and aggressiveness \cite{dorr2014online}. There exist two approaches to personalize ADAS depending on DS:  individual-based  and  group-based.

Individual-based personalization strategies try to reproduce or identify the DS of a given individual using ML techniques or mathematical models. Within this scope, in the works by the authors of \cite{bifulco2008experiments,bifulco2013development}, ACC was adapted to individual drivers in real time based on DS observations, achieved by the recursive least-squares (RLS)-based fitting of a linear car model. The model reproduced the time gaps observed in a short manual-driving session (learning mode) and mimicked these learned time gaps when the personalized ACC was enabled (running mode). In the work by the authors of~\cite{wang2012adaptive}, an ACC was developed with the same  approach as previous, but introducing a forgetting factor with the learning of the driver parameters occurring when the driver is manually controlling the vehicle  while following a lead vehicle. A different approach was chosen in the work by the authors of~\cite{lefevre2015driver}, where a learning, hidden Markov model-based driver model, combined with a model predictive control (MPC) algorithm, was used to create personalized driver assistance able to imitate different DSs. Regarding FCW/AEB, in the work by the authors of \cite{muehlfeld2013statistical}, the personal DS was statistically modeled to estimate  driver-specific probability distribution of danger level to determine the activation threshold of the system. Individual-based strategies have the main advantage of entirely mimicking the DS but, despite this feature being very desirable, it generally requires intensive computation, hard to be achieved in real-time. Moreover, the modeled behaviors would require safety verification since not all drivers handle vehicles in a correct way. To mitigate these drawbacks, group-based approaches have emerged.

Group-based personalization strategies locate drivers in a small number of representative DSs for which a control strategy is implemented. In the work by the authors of \cite{rosenfeld2015learning}, a group-based approach of the driver's ACC preferred time gap is presented. The drivers were clustered to create three general driver profiles to be used, together with demographic information, to predict the gap by using a regression model and decision trees. The authors of \cite{canale2002personalization} describe a stop-and-go ACC system that groups drivers into three clusters depending on their DS, with the cluster membership determining the reference acceleration profile to adjust the ACC controller. Recently, in the work by the authors of~\cite{degelder2016towards}, a support-vector-machine (SVM)-based approach was used to classify driving behavior into two different clusters in order to select a personalization parameter for an ACC. These strategies, despite not entirely mimicking  DS, are computationally more efficient, requiring less computation since they work on a previously offline-trained classification algorithm, allowing online, real-time computation. On the other hand, as they represent a class-averaged DS, they can  easily be validated and verified to always operate in safe margins.

 In this work, a hybrid personalization strategy for DS modeling is proposed. To perform  system development, data from a subset of real-world trips of the Strategic Highway Research Program  (SHRP2)'s naturalistic driving study (NDS) \cite{SHRP2web} are used. First, a group-based technique was used with the aim of building a three-cluster DS classifier. Then, each the clusters was approximated by means of an adaptive neuro-fuzzy inference system (ANFIS) obtaining identification rates higher than 85.7\% for the three clusters. Finally, an individual-based algorithm was used to adapt the behavior of the group to a particular driver. The whole system was  successfully implemented using an FPGA device of the Xilinx’s Zynq programmable system-on-chip (PSoC). The system can mimic the typical timing parameters of a group of drivers as well as tune these typical parameters to model individual DSs. The neuro-fuzzy intelligent sensor provided high speed for real-time active ADAS implementation and could personalize its behavior into safe margins without driver intervention. In~particular, the personalization procedure of the THW parameter for an ACC in steady car-following scenarios~is~described.

The remainder of this paper is organized as follows. Section \ref{sec:description} provides an outline of the proposed DS personalization system. In Section \ref{sec:DSchar}, the driving style characterization methods in car-following scenarios are presented. The neuro-fuzzy modeling approach of driving style groups is provided in Sections \ref{sec:Neurofuzzy} and \ref{sec:implementation} exposes the implementation of the FPGA-based intelligent sensor and provides experiment results for a personalized ACC in a steady car-following situation. Finally,  concluding remarks are summarized in Section \ref{sec:concluding}.

\section{Outline of  Driving-Style Personalization System}\label{sec:description}
\begin{figure}[H]
    \centering
    \includegraphics[scale=0.8]{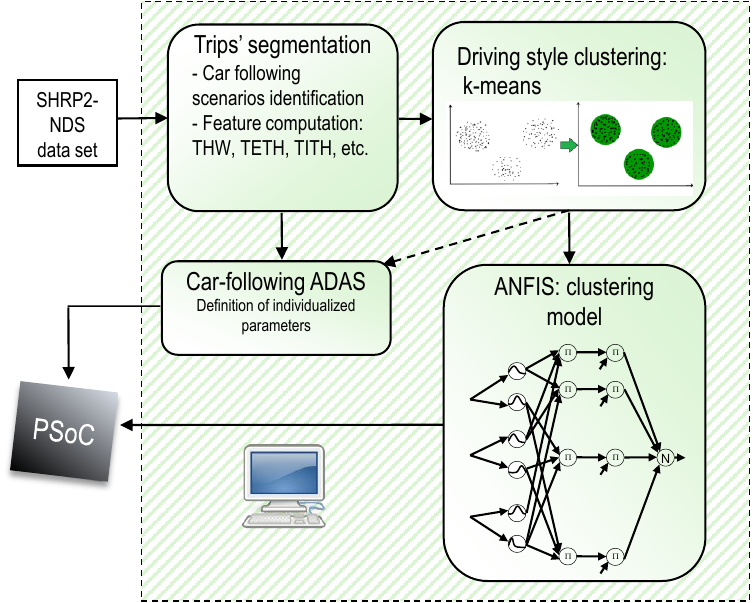}
    \caption{Offline sequence of  tasks involved in the design and development of a neuro-fuzzy sensor for advanced driving-assistance system (ADAS) personalization.}
    \label{fig:figbloq}
\end{figure}

The intelligent sensor developed in this work relies in two well-differentiated stages: an offline design stage and the online in-car operation stage. The sequence of tasks involved in the offline design stage are depicted in Figure \ref{fig:figbloq}. 

The first task comprises the segmentation of  SHRP2-NDS trips with the aim of selecting  car-following scenarios, and the computation of a set of meaningful car-following parameters or features for each segment of the trip (e.g., THW, time-exposed THW (TETH) and time-integrated THW (TITH) \cite{piccinini2014driver}, which will be defined later in Section \ref{sec:drivparams}). Next, an unsupervised clustering technique, the k-means algorithm, was used to group driving styles into a number of clusters in car-following circumstances. Then, an ANFIS was trained in order to develop a high-performance model of the DS classifier. The main advantage of using an ANFIS-like model is that it was suitable for the development of high-speed parallel-hardware architectures, allowing in-car DS classification for ADAS personalization in the online stage. Thus, the ANFIS model was implemented using an FPGA of the Xilinx Zynq device family. More precisely, it is a PSoC that integrates microprocessors and their peripherals with programmable logic.

The FPGA-based driving style classifier acted as an intelligent sensor able to adapt the ADAS response  to  driver preferences in longitudinal car-following scenarios. In particular, a steady car-following application was developed: a personalized ACC.

\begin{figure}[H]
    \centering
    \includegraphics[scale=0.9]{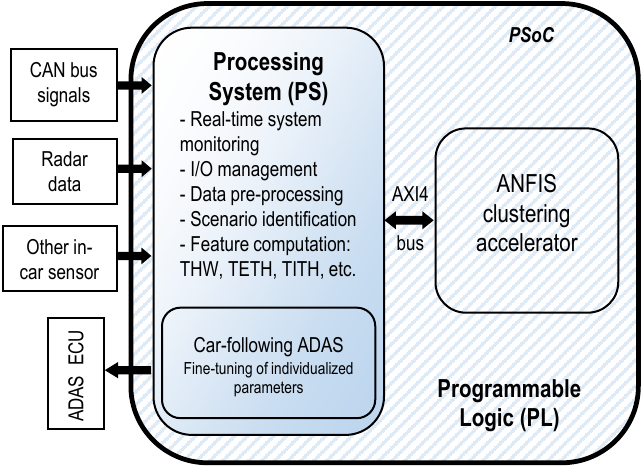}
    \caption{{Block} %Figure moved after the paragraph where it is first mentioned. Please confirm.
    diagram of the field-programmable gate array (FPGA)-based intelligent sensor for online car-following ADAS.}
    \label{fig:figbloq1}
\end{figure}

Figure \ref{fig:figbloq1} depicts a block diagram of the FPGA-based implementation of the intelligent sensor for personalized ADAS. We used a Xilinx Zynq 7000 family PSoC \cite{DS190}. The proposed system was composed of two main modules: the software partition, implemented on the processing system (PS), and the hardware partition, implemented on the programmable-logic (PL) section of the PSoC. %The PS, namely a dual-core ARM Cortex-A9 system, performs global system monitoring, input/output (I/O) management (including the retrieving of vehicle speed and radar data from the field buses of the vehicle) and individualized feature computation for a pre-defined ADAS functionality. A manual driving segment in car-following conditions is required with the aim of updating the current driver driving style. In addition, during the online in-car operation the PS detects car-following scenarios and automatically adapts the parameters of ADAS to the desired driving style.
%On the other hand, the PL comprises typical FPGA resources such as: logic blocks, interconnections, digital signal processing (DSP) cores and random access memory (RAM) blocks. The module implemented in the PL is a high-speed parallel hardware accelerator of the ANFIS model able to provide real-time response even for high-demanding ADAS. The PS/PL communication is performed by means of the standard Advanced eXtensible Interface 4.0 (AXI4) bus. The AXI4 bus enables the PS to share the computed features specified in Figure \ref{fig:figbloq1} with the PL. It also allows the PS to retrieve the identification results provided by the ANFIS clustering accelerator deployed within the PL. Finally, with these identification results, the PS computes individualized ACC parameters and sends them to the car control systems through its field buses. 
Note that the PS consisted of a dual-core ARM Cortex A9 microprocessor, whereas the PL comprised typical FPGA resources such as logic blocks, interconnections, digital signal processing (DSP) cores, and random access memory (RAM) blocks. Additionally, the PSoC system had several interfacing hard modules for field buses (Ethernet, CAN-bus, etc.). Considering  the characteristics of PS and PL, the distribution of the tasks to be performed by the proposed system is explained in the following lines.

The PS, apart from performing global system monitoring, was connected to the vehicle systems  through field buses, managing the I/O interface of the vehicle's systems with the proposed solution  in this work. Therefore, the PS was responsible for capturing the input data from both the radar sensor and the standard information from the CAN-bus. Regarding input data signals, the system is fed with the distance with the preceding vehicle and relative speed between the host vehicle and the preceding one (both from the radar sensor), and with the host vehicle speed (standard information from the CAN-bus). With those signals, the PS computes the driving features that allow to perform  DS classification. These features, stated in Figure \ref{fig:figbloq1}, are computed according to Equations (\ref{eq:THW}), (\ref{eq:TETH}) and (\ref{eq:TITH}) from Section \ref{sec:drivparams}. Inside the PSoC, once the features have been computed, they are sent from the PS to the PL by means of an internal bus, on which the entire PSoC architecture is based: the Advanced eXtensible Interface 4.0 (AXI4-bus). According to features sent from the PS to the PL through the AXI4 bus, the VHDL-based PL-implemented ANFIS clustering accelerator classifies DS and sends the classification results back to the PS through the AXI4-bus. With the received classification results, the PS computes the ACC personalization parameter and, finally, sends it through the vehicle’s CAN-bus to its electronic control unit (ECU), responsible for the ACC.

%Thus, the whole system receives information from the measurement units installed in the car and eventually from a user interface through the CAN-bus or other field buses, computes the features needed for the ANFIS-based clustering, identifies the individual driving style by an FPGA-based ANFIS clustering accelerator, generates the personalized stop-and-go ACC parameters regarding the FPGA-based ANFIS identification and sends those individualized ACC parameters to the ADAS ACC electronic control unit (ECU) of the vehicle. In Section \ref{sec:implementation}, an ACC application with personalized behavior will be presented.

Thus, in summary, and according to Figure \ref{fig:figbloq1}, the PS executes the tasks of interfacing with the vehicle buses to collect the input data, computes  selected identification features, sends them to the ANFIS accelerator deployed in the PL through AXI4 bus, collects the outputs of the accelerator, computes the personalized ACC parameters, and sends them to the ACC module. On the other hand, as it is computationally intensive, the PL implements  an ANFIS-based classification that needs to be executed as fast as possible. Using this architecture, in Section \ref{sec:implementation}, an ACC application with personalized behavior is presented.

\section{Driving Style Characterization in Car-Following Scenarios}
\label{sec:DSchar}
%DS is a very complex concept which depends on multiple factors, forcing multiple terms to arise. Consequently, DS has no agreed definition, but, different authors propose several definitions that match in some of their terms. On the other hand, not only do the motorist's intrinsic factors influence the DS, but also external conditions play an important role \cite{dorr2014online}\cite{gilman2015personalised}. These external conditions modify the way the driver handle the vehicle and range from weather conditions to even the day of the week. Considering applications, there are several fields where DS identification could be of interest, such as drowsiness detection, distraction detection, early-warning applications, driver performance assessment, ride comfort improvement, eco driving, road and vehicle condition monitoring, fleet management, accident detection, insurance applications, hijacking detection and the most important, intelligent vehicle systems and autonomous vehicles.

%\subsection{Driving studies}
Many research works have been conducted in the field of intelligent vehicles, and their sensors, to understand how drivers and traffic behave and to determine which sensors are suitable for each situation. Most of these studies rely on cars instrumented with CAN-bus loggers, inertial measurement units (IMUs), radars, lidars, and video cameras to collect meaningful data. There exist two main branches in driving studies: non-naturalistic and naturalistic.

Non-naturalistic studies, such as NU-Drive \cite{meiring2015review}, UYANIK \cite{abut2009real}, and UTDrive \cite{angkititrakul2007utdrive} make use of dedicated instrumented cars, which simplifies data collection and logistics by increasing the number and complexity of the boarded sensors. The selected human subjects (motorists) do not drive their own cars, so the collected data may not reflect real driving situations. This approach also involves simulator-based road-safety studies, self-report studies, statistical analysis, and authority-investigated crashes. 
However, despite the fact that these methods have greatly contributed to understanding how road users behave and which factors involving crashes are the most important, they do not reflect completely realistic situations. Therefore, NDSs have been conducted to compile data that faithfully reflects driver behavior in every-day traffic situations. NDSs, pioneered by the Virginia Tech Transportation Institute (VTTI) \cite{regan2012naturalistic}, are the most recent trend in traffic safety and ADAS research. 

Meaningful examples of NDSs are UDRIVE \cite{eenink2014udrive}, carried out in the EU, and the 100-Car NDS~\cite{neale2005overview} and SHRP2-NDS \cite{dingus2015naturalistic}, carried out by the VTTI. These studies  focused on collecting data from drivers (human subjects) in their own vehicles and environment in everyday trips without interfering in any  normal behavioral patterns, that is to say, with no experiment control. Thus, NDSs allow for the observance of normal driver situations, providing much better feedback to correctly understand drivers' behavior in normal, unguided traffic situations, as participants do not have the feeling of being involved in an experiment. Consequently, NDSs are more useful to customize real ADAS. In the following, the SHRP2-NDS, used in this work, is briefly introduced.

\subsection{SHRP2-NDS Description}
The main objective of the SHRP2 project is to address the influence of driver performance and behavior on traffic safety. This involves  understanding  the way the driver handles and adapts to a vehicle, roadway characteristics, traffic lights, signs, infrastructure,  and other environmental features. SHRP2-NDS offers two key advantages: detailed and accurate precrash information, including objective information about driving behavior, and exposure information, including the frequency of behaviors in normal driving. To take part in this study, the participants' vehicles are checked for their suitability to fit the systems used in the NDS. Next, while the data acquisition system and instrumentation are installed, the participant serves several driver-assessment tests as well as medical examinations for a total  of 2--3 h.

\begin{comment}

SHRP2-NDS is the largest NDS ever conducted, involving 2360 participants as of the ending moment of the study (September 2012). These participants were recruited in six different locations across the United States of America. Each location hosted 150 to 450 vehicles, and these locations with their coordinating groups were:
\begin{itemize}
    \item \textit{Bloomington, Indiana}---Indiana University, 150 vehicles.
    \item \textit{Central Pennsylvania}---Pennsylvania State University, 150 vehicles.
    \item \textit{Tampa Bay, Florida}---CUBRC and University of South Florida, 441 vehicles.
    \item \textit{Buffalo, New York}---CUBRC, 441 vehicles.
    \item \textit{Durham, North Carolina}---Westat, 300 vehicles.
    \item \textit{Seattle, Washington}---Battelle, 409 vehicles.
\end{itemize}

The study, designed by the VTTI, required the same number of participants for each of age and gender groups. SHRP2-NDS adheres to the principles of informed consent and privacy requirements and each institution operated under the monitoring of either their own Institutional Review Board (IRB) or the VTTI IRB. The participants receive an annual incentive of \$500, and they must authorize  access to the vehicle so as to replace the hard disk in which the data are recorded  every 4--6 months. 
\end{comment}

\begin{figure}[H]
    \centering
    \includegraphics[width=0.9\textwidth]{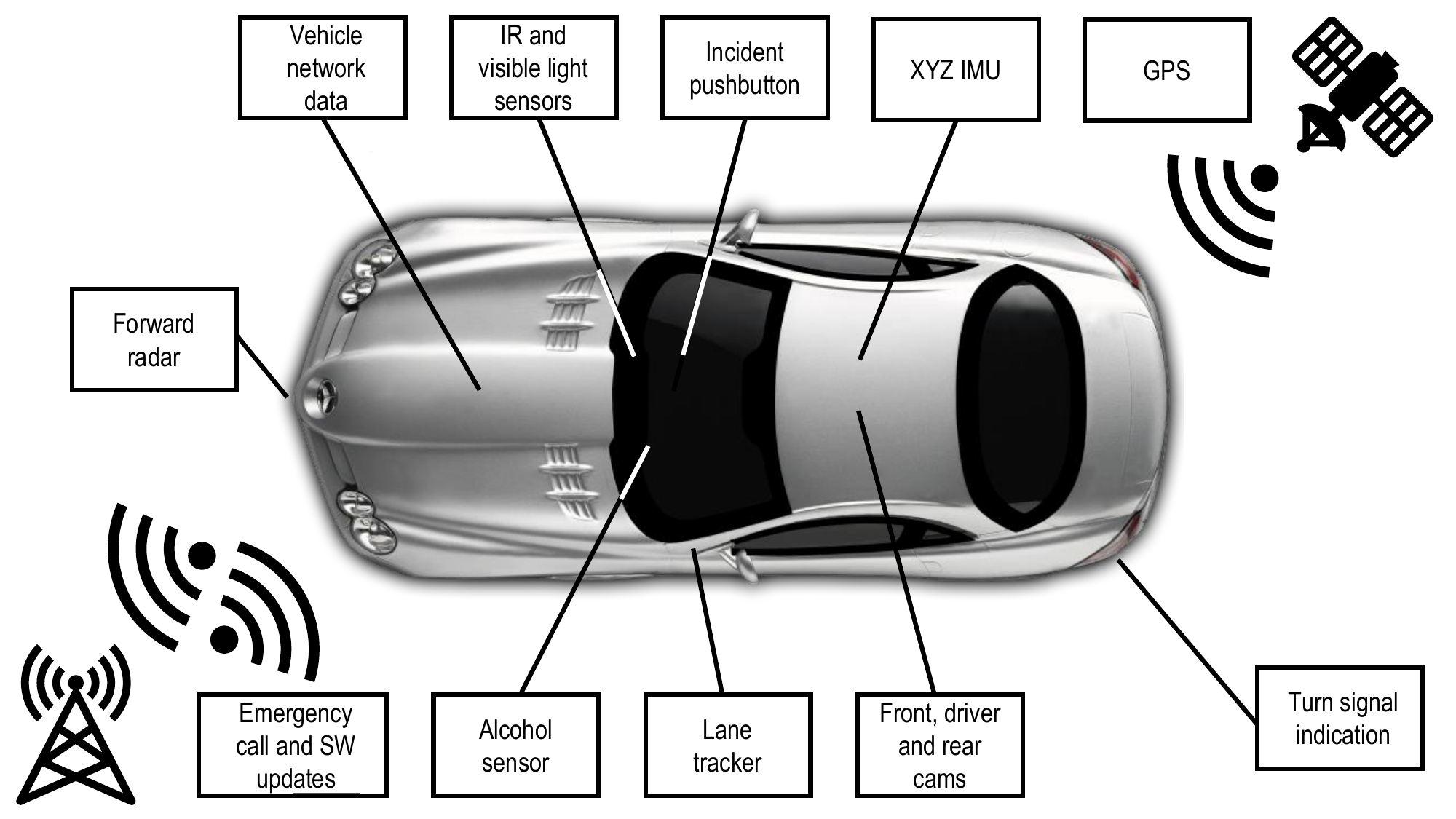}
    \caption{Data acquisition systems and sensors installed in the vehicles that participated in the SHRP2-NDS. IR: infrared; SW: software.}
    \label{fig:figinstru}
\end{figure}

The VTTI developed a custom data acquisition system for the SHRP2-NDS \cite{dingus2015naturalistic}. This system, whose main blocks are displayed in Figure \ref{fig:figinstru}, was manufactured by American Computer Development Inc. and includes a forward radar, four video cameras (with a forward-facing one, color, and wide-angle view), an IMU with XYZ accelerometers and gyroscopes, vehicle network (CAN bus) data logging, GPS-based location, computer vision-based lane tracking, and data-storage capability. Additionally, the data acquisition system has cellular connectivity to provide emergency-call functionalities, system health checks, and software updates. The captured data, including the radar, are uniformly sampled at a rate of 10 Hz, and all the different sources are properly synchronized.

\begin{comment}%Please check if this should be included in the main text.
\begin{itemize}
    \item Multiple videos.
    \item Machine vision.
    \begin{itemize}
        \item Eyes forward monitor.
        \item Lane tracker.
    \end{itemize}
    \item Inertial Measurements Unit (IMU).
    \begin{itemize}
        \item XYZ accelerometer.
        \item XYZ gyro (rates).
    \end{itemize}
    \item GPS.
    \item Forward radar.
    \begin{itemize}
        \item XY positions.
        \item XY speeds
    \end{itemize}
    \item Cellular connectivity.
    \begin{itemize}
        \item eCall functionalities.
        \item System checks, upgrades.
    \end{itemize}
    \item Illuminance sensor.
    \item Infrared illumination.
    \item Passive alcohol sensor.
    \item Incident push button.
    \item Turn signals.
    \item Vehicle network data:
    \begin{itemize}
        \item Accelerator.
        \item Brake pedal activation.
        \item Automatic braking system.
        \item Gear position.
        \item Steering wheel angle.
        \item Speed.
        \item Horn.
        \item Seat belt information.
        \item Airbag deployment.
        \item More variables
    \end{itemize}
\end{itemize}
\end{comment}

\subsection{Car-Following Situations}
Car-following describes a driver following another driver in a traffic stream. For that purpose, drivers operate throttle and brake pedals to maintain a desired range of distance from the preceding vehicle. The main objective of modeling car-following behavior is predicting the following vehicle speed and distance based on  stimuli provided by the preceding vehicle for a set of road  and driver characteristics. A retroactive approach can be applied in this topic, that is, based on DS and speed, predicting the desired distance between the following and the leading vehicle. This is the base of several ADAS, such as ACC, FCW, or AEB.

\subsubsection{Driving Parameters and Driving-Style Characterization}
\label{sec:drivparams}
The main step to correctly identify DSs is determining the adequate variables to provide a robust enough identification. The signal choice depends, however, on the desired application, which is crucial as any further processing of that data will entirely depend on that choice. Therefore, due to the plurality of applications, ranging from ADAS personalization \cite{martinez2015driving} or driving correction for safety and comfort improvement \cite{del2018driving} to fuel economy advice, there is no recommended set of parameters. Thus, for identifying aggressive drivers, high accelerations must be monitored. On the other hand, speed profiles should be monitored to analyze fuel efficiency. Additionally, the acceleration variable is generally combined with brake activation and speed measurements. In other works, pressure on the brake and throttle pedals are used as reliable indicators to identify DSs \cite{murphey2009driver,miyajima2007driver}. Furthermore, the selection of which signals are the most adequate for each application might be guided by the identification or apparition of some circumstances, or the restriction of the identification problem to some contexts or driving events, such as braking, distance-keeping, roundabouts, cornering, lane changes, or even car-following.

In this work, the DS characterization problem was restricted to steady car-following scenarios where clear distance-keeping behavior was observed. In car-following scenarios, the most relevant variables are the speed of the host vehicle ($v$), the relative speed of the host to the leading vehicle ($v_r$), and the distance between host and leading vehicle ($d$). With these variables as a starting point, we derived features to parameterize car-following scenarios such as time headway (THW) and the inverse of time-to-collision (TTCi), which are commonly used.

 \begin{equation}
   THW=\frac{d}{v}
   \label{eq:THW}
  \end{equation}
  
  \begin{equation}
    TTCi=\frac{v_r}{d}.
    \label{eq:TTCi}
  \end{equation}

\begin{figure}[H]
  \centering
  \includegraphics[width=\textwidth]{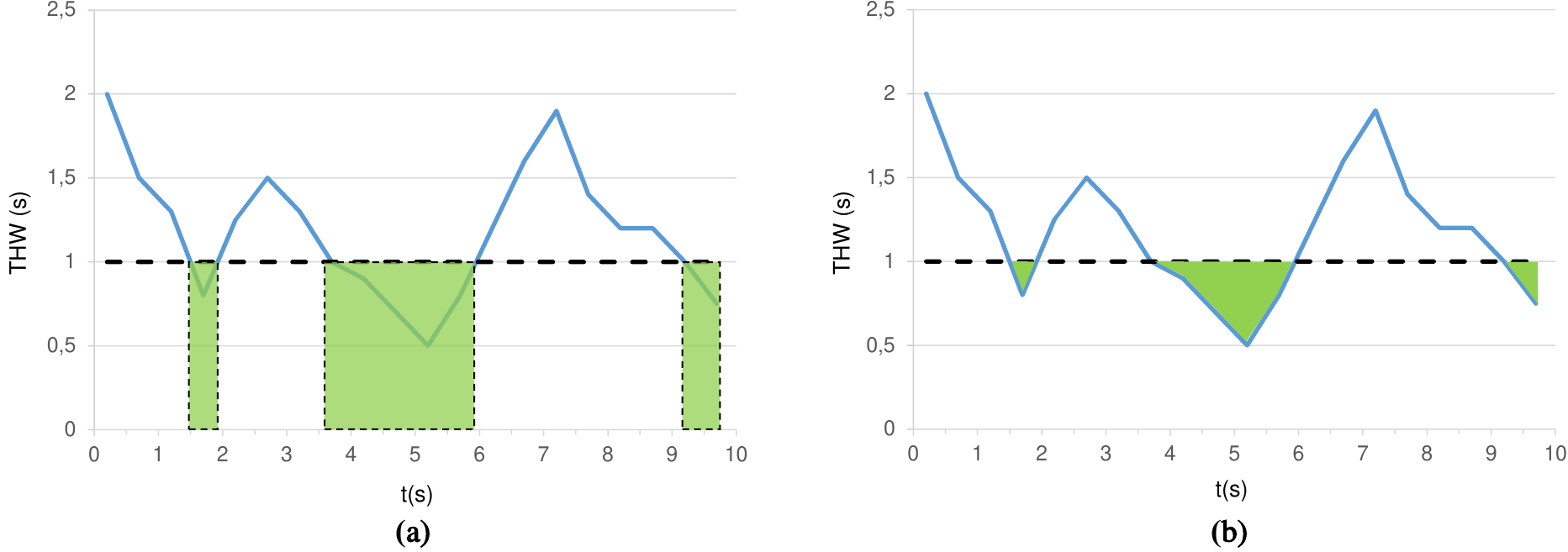}
  \caption{{Representative} example of car-following features: (\textbf{a})  time-exposed time headway (TETH) and (\textbf{b})  time-integrated time headway (TITH).}\label{fig:TETH}
\end{figure}

THW (Equation (\ref{eq:THW})) is the time difference between two successive vehicles when they cross a given point, whereas TTCi (Equation (\ref{eq:TTCi})) is the inverse of the time two vehicles would require to crash if they kept the same speed and trajectory. These parameters are often used to assess car-following styles. Nevertheless, other parameters can be used to provide more complete insight on DS in car-following scenarios. These parameters are time-exposed time headway (TETH) and time-integrated time headway (TITH) \cite{piccinini2014driver}. Thus,  TETH (Equation (\ref{eq:TETH})) represents the time exposure to a THW lower than a predefined safety threshold during a ride.

\begin{equation}
    TETH=\sum_{t\leq T}\delta_i(t) \tau_{s} \qquad \delta_i=\left\{\begin{matrix*}[l]  1\qquad \forall\text{ }0\leq THW_i(t)\leq THW^* \qquad \text{else},\\ 
    0 \end{matrix*} \right.
    \label{eq:TETH}
\end{equation}
where $T$ is the total time interval considered, $\delta_i(t)$ is a binary activation parameter, $\tau_{s}$ is the sampling period, $THW_i(t)$ is the instantaneous value of THW at a given moment t, and $THW^*$ is the predefined safety threshold value (Figure \ref{fig:TETH}a). On the other hand, TITH (Equation (\ref{eq:TITH})), is the summation of the difference between $THW^*$ and $THW_i(t)$ restricted to  time intervals when $THW_i(t)< THW^*$ (Figure~\ref{fig:TETH}b)

\begin{equation}
    TITH=\sum_{t\leq T}\left[ THW^*-THW_i(t)\right]\delta_i(t) \tau_{s} \qquad \delta_i=\left\{\begin{matrix*}[l]  1\qquad \forall\text{ }0\leq THW_i(t)\leq THW^* \qquad \text{else}.\\ 
    0 \end{matrix*} \right.
    \label{eq:TITH}
\end{equation}

The above parameters have been found to characterize the following behavior in both motorists used to activate the ACC mode and non-ACC users \cite{piccinini2014driver}. Thus, as TETH is the time a driver rides behind a leader car below a certain THW threshold, we can determine which percentage of that trip occurs at a time distance below the recommended values. Simultaneously, TITH enables to measure how close a following vehicle has got to its leader during the TETH. Consequently, the use of these parameters jointly with the THW root-mean-square (RMS) value $\text{THW}_\text{RMS}$ provides a good measure of driver behavior. As is seen later, this set of features is very helpful to identify car-following driving~styles.

\subsubsection{Steady Car-Following Premises}
\label{sec:premises}
Car-following scenarios include accelerating, braking, approaching, and steady following \cite{WANG2014159}. Steady car-following circumstances occur when the relative speed between vehicles is low and $\left|{TTCi}\right|\leq0.05$ $\text{s}^{-1}$ \cite{WANG2014159}. The statements used in this work to segment the trips into steady car-following stretches are the same as those used in the work by the authors of \cite{higgs2014segmentation}. Thus, it was assumed that there was a lead vehicle in front of the host vehicle and this leader traveled in the same lane. Additionally, the lead vehicle must have been at a maximum distance of $120 \text{ m}$ and the host vehicle must have been traveling at 20 km/h at least. The maximum distance was  constrained because the radar sensor could detect targets beyond that $120 \text{ m}$ range, and following behaviors with vehicles at such  distance are negligible. On the other hand, a minimum-speed constraint was  applied to filter traffic jams, which cannot be considered real car-following. Additionally, the segments of interest were restricted to those lasting more than 30 s.

\subsection{Car-Following Stretches in the SHRP2-NDS Trips}
The selected car-following stretches were extracted from 48 different trips of 40 different drivers~\cite{VTT1}. Each driver was identified with a numeric code that eased  identification of the driving data  while preserving their privacy. Most of the trips contained mixed-environment driving, ranging from parking lots and streets to motorways and highways. These trips were selected so as to involve different traffic situations as well. Different traffic situations enable researchers to better understand driver behavior and how drivers relate to each other in complex contingencies in both regular transit and safety compromising events. The segmentation of trips into car-following stretches is not trivial, and many parameters should be considered to perform it.

\subsubsection*{Steady Car-Following Segments}
\label{sec:partitioned}
After applying the premises of Section \ref{sec:premises} over the 48 trips, a total of 115 continuous car-following stretches were segmented. Nevertheless, these 115 stretches were extracted from 28 of the 48 selected trips, as the 20 remaining trips did not contain stretches that gathered the characteristics delimited in  steady car-following premises. It is worth noting that the segmented stretches were not evenly distributed among the 28 trips. Thus, in order to uniformize the length of the segmented stretches and consequently reduce  standard deviation to increase comparability, all  stretches were split into smaller ones lasting between 30 and 59.9 s. Additionally, those with ${THW}_{RMS}>4.5\text{ s}$ were discarded because with this THW we could not assure significant car-following events. This new partition was composed of 176 uniformized segments with a duration of $\text{T}_\text{RMS}=37.3$ s and a standard deviation of $\sigma=8.18$ s.

\section{Neuro-Fuzzy Modeling of Driving-Style Clusters}
\label{sec:Neurofuzzy}
The first task involved in the design of the neuro-fuzzy sensor was the segmentation of the SHRP2-NDS trips into the set of steady car-following segments introduced in Section \ref{sec:partitioned} and the computation of the selected features: $\text{THW}_\text{RMS}$ (Equation (\ref{eq:THW})), $\text{TETH}$ (Equation (\ref{eq:TETH})), and $\text{TITH}$ (Equation (\ref{eq:TITH})) for each one of the 176 segments. These parameters are representative of a longitudinal DS in steady car-following situations; therefore, they could be used to personalize ADAS. Consequently, this task consists in grouping together similar driving styles using a clustering approach.

%The following stage of the system is the classification of the segmented steady car-following sub-stretches depending on their characteristics, this is, DS classification. There exists a large number of applications and classification criteria, and, according to these criteria, there are many approaches for DS classification. A group-based DS identification approach is used in this work, and for this approach, clustering techniques are the most appropriate.

\begin{comment}
Grouping driver styles into discrete classes is the general tendency. These classes are usually based on the distribution of  selected driving parameters and extracted features. To carry out this classification task,  driving classes must be defined in advance according to the characteristics of the available data. Additionally, a trade-off exercise between classification accuracy and complexity must be put into practice.

Clustering techniques are data-mining techniques used to place data elements into their related groups and consequently allow to perform this grouping, and to analyze the trade-off of how fine the groups are and how complex the identification system will be. Clustering techniques can be divided into two families: univariate and multivariate clustering. In the case of DS identification, since multidimensional complex data were used, multivariate clustering operations were performed.
\end{comment}

\subsection{Driving-Style Clustering}
Several clustering algorithms have been used to distinguish DSs and DS-class labeling \cite{castignani2015driver}. The~selection of a concrete algorithm depends on the trade-off between complexity and performance.  Mean-shift clustering is based on finding high-density data areas by means of a sliding window of a specified radius, aiming to locate the centroid of each area. This algorithm has the advantage of not needing to know the number of desired clusters, as the algorithm detects them by itself, but as a weakness; it should be pointed out that the selection of the window radius may be nontrivial. Another used clustering algorithm is density-based spatial clustering of applications with noise (DBSCAN)~\cite{shen2016real}, which can filter  outliers and find arbitrarily shaped and sized clusters, but does not perform well when clusters have variable density. Expectation--maximization (EM) clustering using Gaussian mixture models (GMM) is a flexible algorithm in terms of cluster covariance \cite{moon1996expectation}, which bridges the restriction of distance-based solutions that only work on circular-shaped clusters. Additionally, since GMMs use a probability cluster, a given datapoint can belong to several data clusters with different probability, allowing mixed membership. Agglomerative hierarchical clustering (HCA) relies on building a tree depending on the similarity/dissimilarity between data points/clusters \cite{zhao2002evaluation}, and~due to this characteristic, HCA does not need to preselect a number of clusters and it is particularly suitable to recover underlying hierarchical data structures. However, when there are particularities in some of the input data, HCA tends to group all of those particular points together, causing cluster unbalance. Finally, several research pieces   in driver identification \cite{kalsoom2013clustering} and road condition monitoring~\cite{bhoraskar2012wolverine} have successfully used the k-means algorithm, as it useful for the proposed group-based DS identification application. The k-means algorithm is simple and quick since it is based on computing  distances between each point and the groups' centroids. However, since the number of clusters must be previously specified, and the centroids of each cluster are randomly initialized, the repeatability of this type of clustering is not always assured. Nevertheless, it is quick enough to execute multiple runs in a reasonable period of time. Additionally, input data groups elaborated by k-means can  easily be interpreted. Hence, due to  prior characteristics, the k-means algorithm has been used in this work to carry out DS grouping.

\subsubsection*{K-Means Clustering Results}
\label{sec:clustering_results}

\begin{figure}[H]
    \centering
    \includegraphics[scale=0.5]{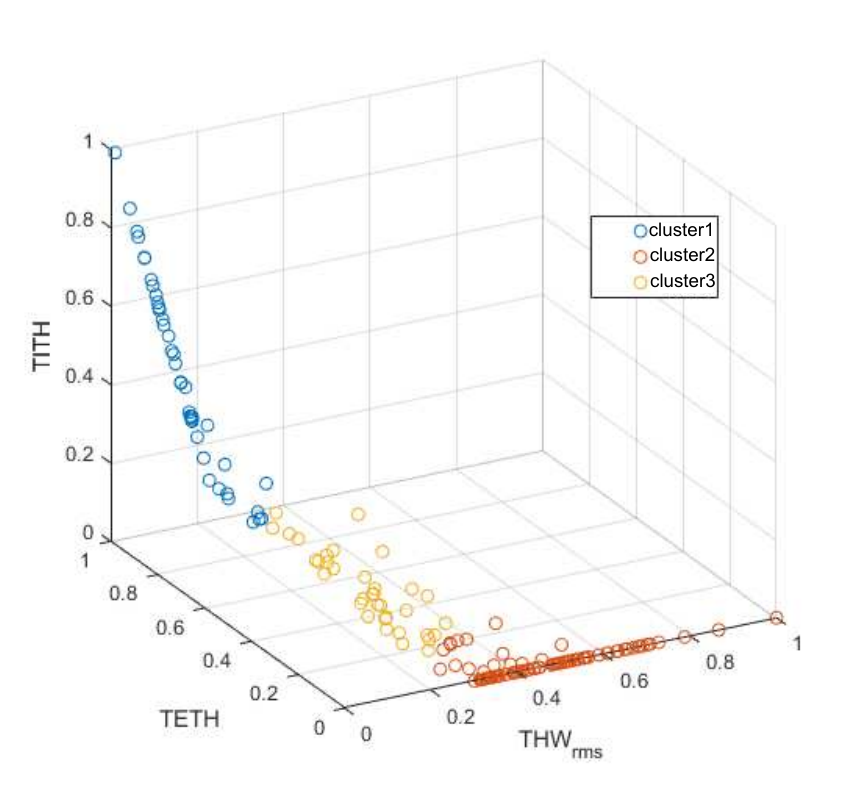}
    \caption{Clusters obtained applying the k-means algorithm to the car-following segments; $\text{THW}_\text{RMS}$, $\text{TETH}$, and $\text{TITH}$ values were normalized.}
    \label{fig:figclus}
\end{figure}

First,  the selected features $\text{THW}_\text{RMS}$, $\text{TETH}$, and $\text{TITH}$ features were computed for each of the 176 driving segments and normalized into the [0,1] range.  According to the minimum following safety threshold of \cite{fleming2018adaptive} and the selected THW in the work by the authors of \cite{piccinini2014driver}, TETH and TITH were calculated for a critical value of $\text{THW}^*=1.5$ s. After that, three-group k-means clustering of those segments was performed. The obtained cluster structure is depicted in Figure \ref{fig:figclus}. Note that, for $\text{THW}_\text{RMS}$ values higher than $\text{THW}^*$ 1.5 s, TETH and TITH were always 0, generating the blank zone at the right $\text{TETH}-\text{THW}_\text{RMS}$ semiplane of the figure.

\begin{figure}[H]
\centering{ \includegraphics[width=0.89\textwidth]{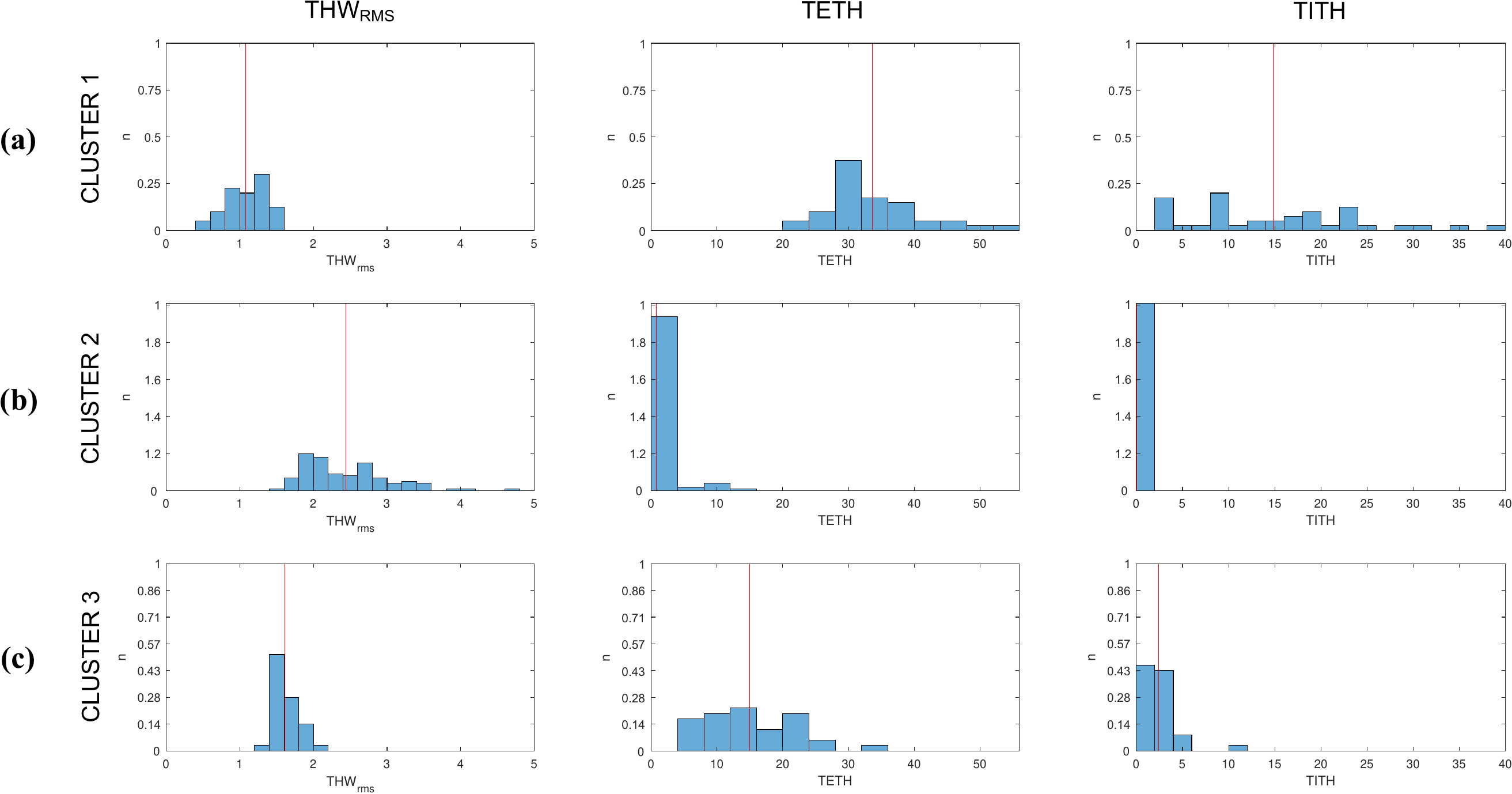}}
 \caption{Histogram of $\text{THW}_\text{RMS}$, $\text{TETH}$, and $\text{TITH}$ values distribution for  (\textbf{a}) Cluster 1,  (\textbf{b}) Cluster 2, and (\textbf{c}) Cluster 3. Red line represents  average value of each distribution.}
 \label{fig:barras}
 \end{figure}

  DS groups were  found to be stable and highly reproducible despite the randomness of the cluster centroid initialization. Therefore, within these data, a unique solid structure could be found. In~Figure~\ref{fig:barras}, the distribution of $\text{THW}_\text{RMS}$, $\text{TETH}$, and $\text{TITH}$ values according to this normalized cluster structure is shown. %Thus, according to the clustered data representation, the three identified clusters have the following characteristics:
 Given the distributions displayed in the figure, the clusters could be described as follows:
 
 \begin{itemize}[leftmargin=*,labelsep=5.8mm]
    \item Cluster 1: groups the drivers with the lowest $\text{THW}_\text{RMS}$ and the highest $\text{TETH}$ and $\text{TITH}$. This cluster is representative of the most aggressive car followers.
        \item Cluster 2: groups the drivers with high $\text{THW}_\text{RMS}$ values and minimum $\text{TETH}$ and $\text{TITH}$. Thus, it~incorporates the least aggressive car followers.
    \item Cluster 3: groups the drivers with low $\text{THW}_\text{RMS}$ values, medium to low $\text{TETH}$ and the lowest $\text{TITH}$, representing medium aggressive car followers.

\end{itemize}

%\subsection{Clustering execution}
%\subsection{Description of the final clusters}

\subsection{ANFIS-Based Identification}

The second task  in the development of the proposed intelligent sensor was the development of a high-performance model of the DS classifier obtained above. AS the proposed system was intended to accomplish online DS identification, clustering techniques could not be directly used. There is a variety of solutions suitable for efficient real-time hardware implementation. Thus, artificial neural networks (ANNs) were  used in the work by the authors of \cite{brombacher2017driving} to score drivers depending on the safety of their DS. In the work by the authors of \cite{kurt2011probabilistic}, finite-state machines (FSMs) were  used to decide whether a driver belongs to a DS class depending on their driving decisions. Nevertheless, an outcoming line of work in DS identification is based on fuzzy-logic implementations \cite{zadeh1988fuzzy}. Thus, in the work by the authors of \cite{aljaafreh2012driving}, anomalous DS was identified applying a fuzzy inference system (FIS) on accelerometer data, and an FIS-based dangerous DS identification application was proposed in the work by the authors of \cite{choudhary2014smart}; in the work by the authors of \cite{dorr2014online}, fuzzy logic was applied to identify DSs in an online~fashion.

To accomplish this task, an adaptive neuro-FIS (ANFIS) was used since this system was suitable to model clusters with online performance. Thus, once the k-means clustering algorithm classified the steady car-following segments of Section \ref{sec:partitioned} into three DS clusters depending on their $\text{THW}_\text{RMS}$, $\text{TETH}$, and $\text{TITH}$, an ANFIS was trained for each one of the three clusters. Each ANFIS model returned a continuous value indicating the fitting of the prior input parameters into each of the clusters. Attending to those output values, the ANFIS model with the maximum output identified the cluster to which the inputs belong.

%\subsubsection{Membership Functions Definition}

%For each of the input variables, 3 membership functions are elaborated depending on their linguistic values, namely the $LOW$, $MED$ and $HIGH$ functions. These $A_{ij}$ bell-shaped membership functions are defined by Equation \ref{eq:gbell}

%\begin{equation}
 %   \mu_{ij}(x;a,b,c)=\frac{1}{1+\left|\frac{x-c}{a}\right|^{2b}}\qquad \text{with}\text{  }i=1,\dots,m\text{  }\text{and}\text{  }j=1,\dots,n, 
  %  \label{eq:gbell}
%\end{equation}

%where $m$ is the number of membership functions for each input, $n$ is the number of inputs, $a$ defines the width of the membership function, $b$, the steepness of the curves at each side of the center plateau, and $c$, the center of the function. %In this particular application, since there are 3 membership functions defined for each of the 3 inputs, $n=m=3$. %The sets of membership functions are named $A_{i1}$ for the $THW_{rms}$ set, $A_{i2}$ for the $TETH$ set and $A_{i3}$ for the $TITH$ set.

\begin{comment}

\begin{figure}[H]
    \centering
    \includegraphics[width=0.75\textwidth]{THW_member.PNG}
    \caption{Caption}
    \label{fig:THW}
\end{figure}

\begin{figure}[H]
    \centering
    \includegraphics[width=0.75\textwidth]{TETH_member.PNG}
    \caption{Caption}
    \label{fig:TETH}
\end{figure}

\begin{figure}[H]
    \centering
    \includegraphics[width=0.75\textwidth]{TITH_member.PNG}
    \caption{Caption}
    \label{fig:TITH}
\end{figure}

\end{comment}

\subsubsection{Zero-Order Takagi--Sugeno Inference System}
The ANFIS model used in this work was based on a zero-order Takagi--Sugeno inference system using prod-sum operators with the product for the inference process and the sum for the aggregation of the rules \cite{jang1993anfis,jang1995neuro}. Consider set of rules:

\begin{equation}
    {R}_{j}: \text{IF }x_1\text{ IS }A_{1j}(x_1)\text{ AND }x_2\text{ IS }A_{2j}(x_2)\text{ AND }\cdots x_n\text{ IS }A_{nj}(x_n)\text{ THEN }y\text{ IS }c_j,
\end{equation}
where $R_j$ is the $j$th rule $(1\leq j\leq m)$, $x_i(1\leq i \leq n)$ are input variables, $y$ is the output, $c_j$ is a constant consequent, and $A_{ij}(x_i)$ are linguistic labels; each one is associated with a membership function, $\mu_{ij}(x_i)$. In zero-order Takagi--Sugeno fuzzy models, the inference procedure used to derive the conclusion for a specific input $\mathbf{x}=\left(x_1^0,x_2^0,\cdots,x_n^0\right)$ consists of two main steps. First, the firing strength or weight $w_j$ of each rule is calculated as

\begin{equation}
    w_j=\prod_{i=1}^n \mu_{ij} \left(x_i^0\right).
    \label{eq:strength}
\end{equation}

Next,  overall inference result $y$ is obtained by means of the weighted average of the consequents

\begin{equation}
    y=\dfrac{\sum_{j=1}^m w_j c_j}{\sum_{j=1}^m w_j}=\dfrac{N}{D}.
    \label{eq:N/D}
\end{equation}

In this particular application, there are $n=3$ inputs ($x_1=\text{THW}_\text{RMS}$, $x_2=\text{TETH}$, and $x_3=\text{TITH}$), whereas for each of the inputs, three linguistic labels, namely, $LOW$, $MEDIUM$, and $HIGH$, were  assigned and 27 rules were generated ($m=27$). Membership functions $\mu_{ij}$ associated to these labels are bell-shaped functions, which are defined as follows,

\begin{equation}
    \mu_{ij}(x;a,b,e)=\frac{1}{1+\left|\frac{x-e}{a}\right|^{2b}}, 
    \label{eq:gbell}
\end{equation}
where $a$ defines the width of the membership function, $b$ is the steepness of the curves at each side of the center plateau, and $e$ is the center of the function.

Finally, to train the ANFIS, a combination of a least-squares estimator (LSE) and a gradient-descent method (GDM) were used. Each epoch of this learning process was composed of a forward pass and a backward pass. In the forward pass,  consequent parameters, $c_j$, were identified by the LSE method, and in the backward pass,  antecedent parameters $a$, $b$, and $e$ were updated by the GDM.

\subsubsection{ANFIS Training}
\label{sec:ANFIStrain}

The 176 segments from Section \ref{sec:partitioned} were labeled in accordance with the clustering described in Section \ref{sec:clustering_results} (see Figure \ref{fig:figclus}). Once the segments were labeled, they were partitioned into a training set (75\% of the samples) and a test set (25\% of the samples). Each ANFIS cluster was trained and tested with the same set of data since they were designed to decide whether the input data belong to the cluster they represent. Consequently,  membership was represented with a value of ``1'' if the data belonged to the cluster modeled by that ANFIS, and with ``0'' if not.

%Initially, in the training process, the parameters of the membership functions ($\mu_{ij}$) are kept fixed and the consequent values $c_j$ are tuned by means of a least-squares method. Once the initial tuning is finished, the error signals are back-propagated so as to update the $\mu_{ij}$ parameters by a gradient-descent algorithm while keeping $c_j$ constant.

\begin{figure}[H]
    \centering
    \includegraphics[width=\textwidth]{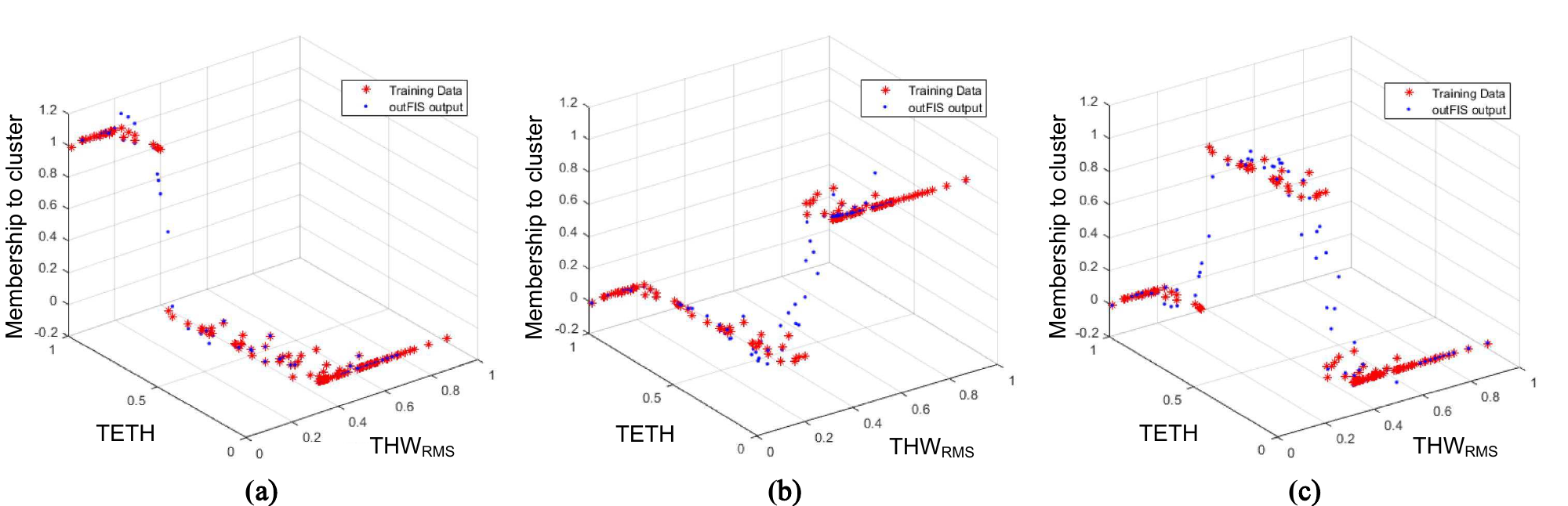}
    \caption{ (\textbf{a}) Adaptive neuro-fuzzy inference system (ANFIS) 1 (Cluster 1), (\textbf{b}) ANFIS 2 (Cluster 2), and (\textbf{c}) ANFIS 3 (Cluster 3). Training data  shown in red; response of  corresponding trained ANFIS  shown in blue.}
    \label{fig:ANFIStrain}
\end{figure}

As can be seen in Figure \ref{fig:ANFIStrain},  $\text{THW}_\text{RMS}$ and $\text{TETH}$ values were the same for all the clusters, and despite this figure being a 3D plot and $\text{TITH}$ not being able to be displayed, it also coincided for all three ANFIS models. It could also be observed that the value of the Z-axis was ``1'' when  data points belonged to the cluster to be modeled, and ``0'' when not.

\subsubsection{ANFIS Testing and Identification Results}

After the training stage, the remaining 25\% of the steady car-following segments (see Section~\ref{sec:partitioned}) are used to test the identification performance of the ANFIS-based DS identificator. Thus, the test segments were  simultaneously input to the three ANFIS-clusters, and the neuro-fuzzy system with the highest output was considered to be the class to which the segment belonged. With this procedure, the~outputs of the system were evaluated, showing an accuracy mark of $95.45\%$. This mark was much higher and the classification more detailed than those obtained in previous works from other authors, such as the work by the authors of \cite{degelder2016towards}, where, using an SVM, a given driver's DS was classified between only two clusters with an accuracy of  85\%. 

Additionally, the confusion matrix in Table \ref{tab:confusion} gives  deeper insight into this accuracy result. By~analyzing this matrix, accuracy rates of $85.71\text{\%}$ for Cluster 1, $96.43\text{\%}$ for Cluster 2, and $100\text{\%}$ for Cluster 3 were reached by the final ANFIS. Nevertheless,  confusions reflected in the matrix happened between contiguous clusters; hence,  no erratic classification behavior happened while identifying DSs with this~system.

\begin{table}[H]
\caption{Confusion matrix of  ANFIS-based DS identifier.}
\label{tab:confusion}
\centering
\begin{tabular}{cccc}
\toprule
\textbf{Actual/Identified }	& \textbf{Cluster 1}	& \textbf{Cluster 2} &\textbf{Cluster 3}\\
\midrule
\textbf{{Cluster 1}}	% Bold is necessary
& 6	& 0 & 0\\
\textbf{{Cluster 2}}	& 0	& 27 & 0\\
\textbf{Cluster 3}	& 1	& 1 & 9\\
\bottomrule
\end{tabular}
\end{table}

\section{Implementation of  FPGA-Based Intelligent Sensor}

\label{sec:implementation}

A block diagram of the proposed DS-based ADAS personalization solution is  displayed in Figure \ref{fig:figbloq1}. It is a hybrid hardware/software (HW/SW) architecture implemented on the Xilinx XC7Z045-2FFG900 PSoC \cite{DS190} using the Xilinx ZC706 development board \cite{UG954}. The entire HW partition of the system (deployed in the PL of the PSoC) was implemented using VHDL language and the Xilinx Vivado 2018.1 design suite. On the other hand, the remainder of the proposed system with its functionalities was to be programmed at the PS by developing a bare-metal C application that can acquire  vehicle bus data; compute the THW, TETH, and TITH features; share them with the PL; retrieve the ANFIS accelerators' results; compute the personalization parameters; and send them to the ACC ECU.

\subsection{Hardware Partition: ANFIS Accelerators}
%The hardware partition's top-bottom hierarchy, deployed within the PSoC's FPGA can be described as follows:

\begin{figure}[H]
    \centering
    \includegraphics[width=1\textwidth]{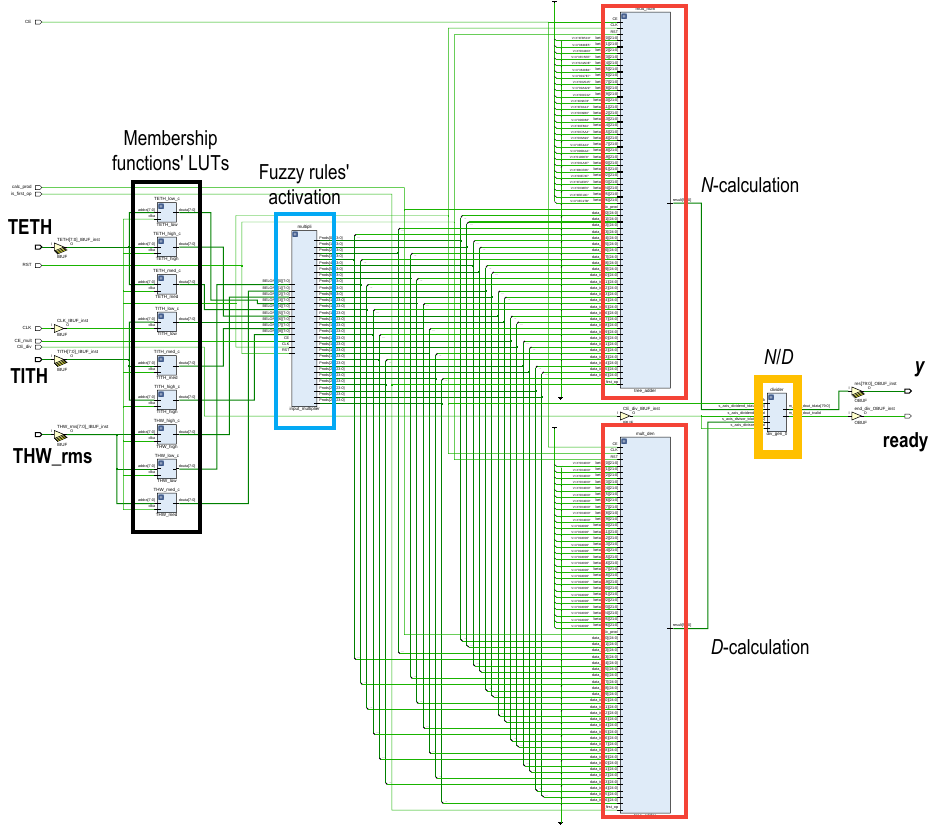}
    \caption{Block scheme of the parallel architecture of a three-input ANFIS implemented in the programmable logic (PL) of the programmable system-on-chip (PSoC). Three ANFIS cores, one per cluster, were implemented in the HW partition.}
    \label{fig:HW_struc}
\end{figure}

The building blocks on which the ANFIS HW accelerators rely were structured according to the Takagi--Sugeno Inference System (Equations (\ref{eq:strength})--(\ref{eq:gbell})). After several tests,  system inputs ($\text{THW}_\text{RMS}^0$, $\text{TETH}^0$ and $\text{TITH}^0$) were represented using an 8-bit fixed-point fractional data format, the bit widths of the intermediate operations were properly propagated and trimmed for not losing precision, and  output $y$ was trimmed to a 32-bit two-complement fixed-point representation, chosen to match with the AXI4 bus width. As can be seen in Figure \ref{fig:HW_struc}, the proposed ANFIS architecture was organized in four layers. In the first layer, the membership of the system inputs to the antecedents of the rules were evaluated. Then, in the second layer,  rule activations were concurrently computed (Equation (\ref{eq:strength})). Next, in Layers 3 and 4, the weighted average of the consequents was calculated (Equation (\ref{eq:N/D})). The HW partition was composed of three ANFIS accelerators, one per cluster.%the sum inputs are evaluated by their membership functions and the fuzzy rules are calculated; in the intermediate step, the sum of the elements of the fuzzy set multiplied by their consequents, as well as the sum of the elements of the fuzzy set are computed; and finally, the ANFIS output is calculated by dividing the previously indicated sums in their appearance order. The sum and sum of products have been substituted by accumulations and multiplication-accumulations, respectively, in order to provide scalability and high timing performance.

\subsubsection{Membership Function Evaluation and Fuzzy-Rule Computation}

The generalized bell-shaped membership functions were precalculated and stored as look-up tables (LUTs) (remarked in black in Figure \ref{fig:HW_struc}) at the PL block RAMs (BRAMs). Therefore,  evaluation of the input membership to each antecedent was straightforwardly obtained by addressing those values, lasting only one clock cycle. Once the input membership functions were evaluated, fuzzy-rule activations were calculated. As there were three membership functions for each of the three inputs, 27~weights were to be computed. These fuzzy-rule activations were three-input products, computed by a  fuzzy-rule activation module remarked in blue in Figure \ref{fig:HW_struc}. These products were  efficiently computed by a full-VHDL design intended to only use Xilinx DSP resources \cite{UG479}, improving timing performance. To achieve this DSP-only implementation, the three-input products were done two by two. Thus, first the product of two of the inputs was calculated and stored in an intermediate result pipeline register and second, the stored partial product was multiplied by the remaining input and saved in the output register. This product pipeline required two clock cycles.

\subsubsection{Computation of  Sum and  Weighted Sum of Rule Activation}

\begin{figure}[H]
    \centering
    \includegraphics[width=0.6\textwidth]{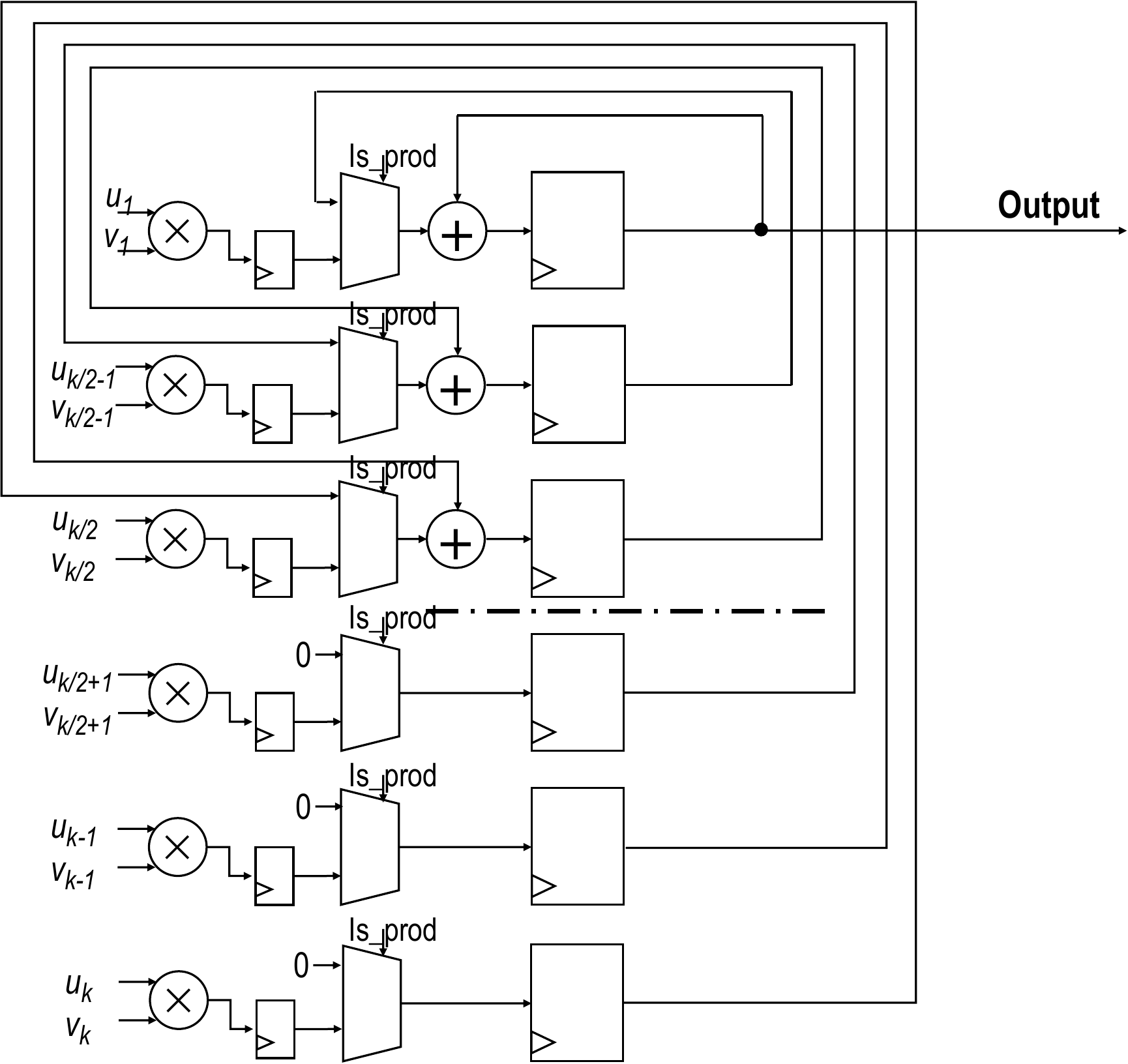}
    \caption{Scheme of  proposed sum of product architecture that substitutes  traditional tree-adder~solution.}
    \label{fig:tree}
\end{figure}

In the work by the authors of \cite{mata2019hardware}, a high-performance product--architecture sum, developed by the authors, was described. This topology, shown in Figure \ref{fig:tree}, replaced the tree-adder architecture. It was intended to minimize latency, save resources, and minimize the number of used DSPs. Thus, for a given number $k$ of products, the proposed architecture only used $k$ multiplier/adder blocks, while a tree adder would spend $2k-1$ of the same hardware resources. This architecture, with inputs $\mathbf{u}=(u_1,\cdots,u_k)$ and $\mathbf{v}=(v_1,\cdots,v_k)$, control signals \texttt{is\_prod} and \texttt{CE}, and output $p$ operates~as~follows.

\begin{enumerate}[leftmargin=*,labelsep=5.8mm]
    \item Product signal \texttt{is\_prod}  set to ``1'' and all  registers are reset.
    \item Product $u_j\cdot v_j$ with $j=1,...,k$  computed and stored in each of the $k$ accumulator registers.
    \item Signal \texttt{is\_prod}  set back to ``0'' and  first accumulation is performed. Thus, accumulator registers from $1$ to $k/2$ contain the sum of $u_j\cdot v_j$ products. Registers from $k/2+1$ to $k$ are now filled with~zeros.
    \item Successive $\left \lceil log_2 k \right \rceil$ accumulations are performed until  valid result is present in register 0.
\end{enumerate}

Two instances of this core, labeled $D$ and $N$ in Figure \ref{fig:HW_struc}, are used to perform the computation of the sum and the weighted sum of rules' activation parallelly ($D$ and $N$ in Equation (\ref{eq:N/D}), respectively). In both modules, $k$ equals the number of rules, that is, $k=27$.  For the $N$-module, $u_j=w_j$ and $v_j=c_j$, whereas for the $D$-module, $u_j=w_j$ and $v_j=1$. A ROM storing the values of $c_j$ is connected to the $N$-module. The latency of this architecture is $\left \lceil log_2 k \right \rceil+2$; thus, with $k=27$, the latency of both instances is seven clock cycles.

\subsubsection{Divider Module}
This is the last layer of the ANFIS accelerator. The divider module was  elaborated by means of the Xilinx IPCore divider generator \cite{PG151}. This IPCore was parameterized to match with the size of the $N$ and $D$ operands using the high-radix division implementation. This particular implementation could be pipelined to achieve good time performance and, as it depends on multiply--accumulate operations, it is optimally deployed in DSP blocks. With the selected word lengths and pipelined, this module requires 43 clock cycles to return valid results.

\subsubsection{Parameterization and Control Signals}
\begin{figure}[H]
\centering{ \includegraphics[width=1\textwidth, height=0.35\textwidth]{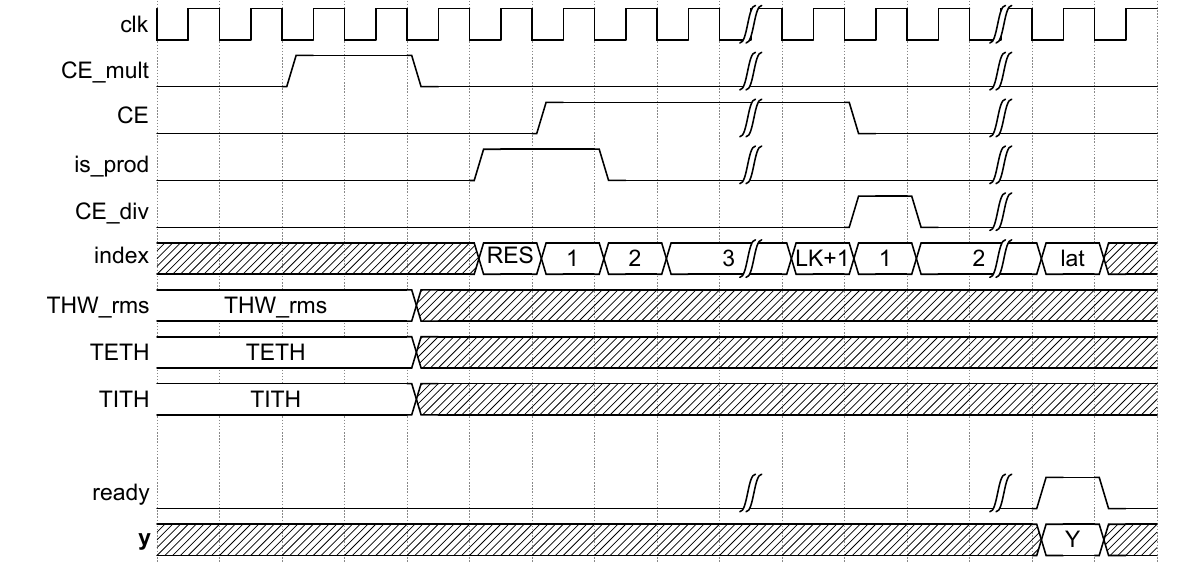}}
 \caption{Chronogram of  control-signal sequence of the ANFIS core.}
 \label{fig:chronogram}
 \end{figure}

The complete structure of the ANFIS HW accelerator is parametric and fully customizable. LUT ROMs containing  membership functions as well as  consequents were simultaneously initialized. Elements such as type depths, signal bit-widths, and number of inputs, number of membership functions, or number of fuzzy rules were defined on a standalone package. The complete ANFIS was controlled by the sequence of control signals represented in the chronogram of Figure \ref{fig:chronogram}.

Control signals of the ANFIS HW accelerator were \texttt{rst}, \texttt{CE\_mult}, \texttt{CE}, \texttt{is\_prod}, and \texttt{CE\_div} (see Figure \ref{fig:chronogram}); they worked as follows.

\begin{enumerate}[leftmargin=*,labelsep=5.8mm]
    \item \texttt{rst} clears pipeline registers and  multiplication--accumulation units.
    \item \texttt{CE\_mult} drives  multipliers of  fuzzy-rule calculation.
    \item \texttt{CE} activates multiplication--accumulation units to iteratively compute $N$ and $D$.
    \item \texttt{is\_prod}, in conjunction with the first cycle of \texttt{CE}, is used to indicate that the multiplier--accumulation unit must store the products of the fuzzy rules by their corresponding consequents instead of performing any accumulation.
    \item \texttt{CE\_div} triggers  divider module calculating the output result of the ANFIS.
\end{enumerate}

Finishing with the  ANFIS HW accelerator implementation, it is worth nothing that the three clusters of Section \ref{sec:clustering_results} must be modeled by this method. Consequently, three instances of this HW accelerator had to be deployed in the PL, each one configured with the parameters the ANFIS cluster to which it corresponds. 

\subsection{Experiment Results}
The three ANFIS cluster HW accelerators were implemented in the selected PSoC, achieving the subsequent results.
\subsubsection{Resource Usage}
The full HW system was successfully implemented, with the postimplementation results displayed in Table \ref{tab:implementation}. The three-ANFIS system fit into the selected PSoC's logic, leaving enough resources available for further system applications, escalations, or improvements.

\begin{table}[H]
\caption{Postimplementation resources report (Xilinx XC7Z045-2FFG900).}
\label{tab:implementation}
\centering
\begin{tabular}{cccc}
\toprule
\textbf{Resource}	& \textbf{Utilization}	& \textbf{Available} & \textbf{\% Used}\\
\midrule
LUT	& 13500		& 218600 & 6.17\\
Flip-flops	& 15759		& 437200 & 3.60\\
RAM	blocks & 15			& 545 & 2.76\\
DSP	& 294			& 900 & 32.76\\
\bottomrule
\end{tabular}
\end{table}

\subsubsection{Timing Performance}

With respect to timing, the three-ANFIS' postsynthesis timing report pointed out that the minimal clock period suitable for application to the designed hardware was 7.122 ns, or a maximal clock frequency of 140.41 MHz. With this maximal clock frequency, the designed HW implementation could be used as an AXI4 peripheral dependent of an AXI4 bus clock frequency of 100 MHz. This design delayed 53 clock cycles (530 ns at $\text{F}_\text{CLK}=100\text{ MHz}$) to return the computed outputs. These results outperformed the timing obtained for the full-software PC-based (20-core Intel Xeon CPU E5-2630 v4 $@$ 2.20 GHz with 32 GB of DD4 RAM) MATLAB model design, with top performance peaks of 1.829 ms to compute the same set of 3 ANFIS, as well as a PC-based, C-coded prototype that achieved timing marks of 12.45 $\upmu$s. 

The obtained timing was better than in other FPGA-based ANFIS approaches, such as the work by the authors of \cite{saldana2016digital}, where timings of \textasciitilde12 $\upmu$s were obtained in the computation of a system with the same number of inputs and outputs (three and one, respectively) as that developed in this work. On~the other hand, in the work by the authors of \cite{darvill2017novelanfis}, a novel ANFIS HW architecture, able to reduce the timing mark of 530 ns achieved in the present work more than  50\%, was presented. Recently, several innovative architectures on other ML algorithms have been proposed with the aim of achieving extreme timing performance results. Examples of these innovations are a HW implementation of a radial basis function (RBF) network, for which operational frequencies of up to 450 MHz for high bit-width inputs were achieved \cite{fernandes1}, and an SVM  implementation able to be run up to 20 times faster than other state-of-the-art techniques \cite{fernandes3}. 

Consequently, the hybrid HW/SW implementation developed in this work is an innovative solution between conventional SW-based approaches and novel FPGA-based, extreme performance architectures, which, provides an adequate trade-off between complexity, performance, and~development~time.

\subsubsection{ACC Personalization Application}

\begin{figure}[H]
\centering{ \includegraphics[scale=0.7]{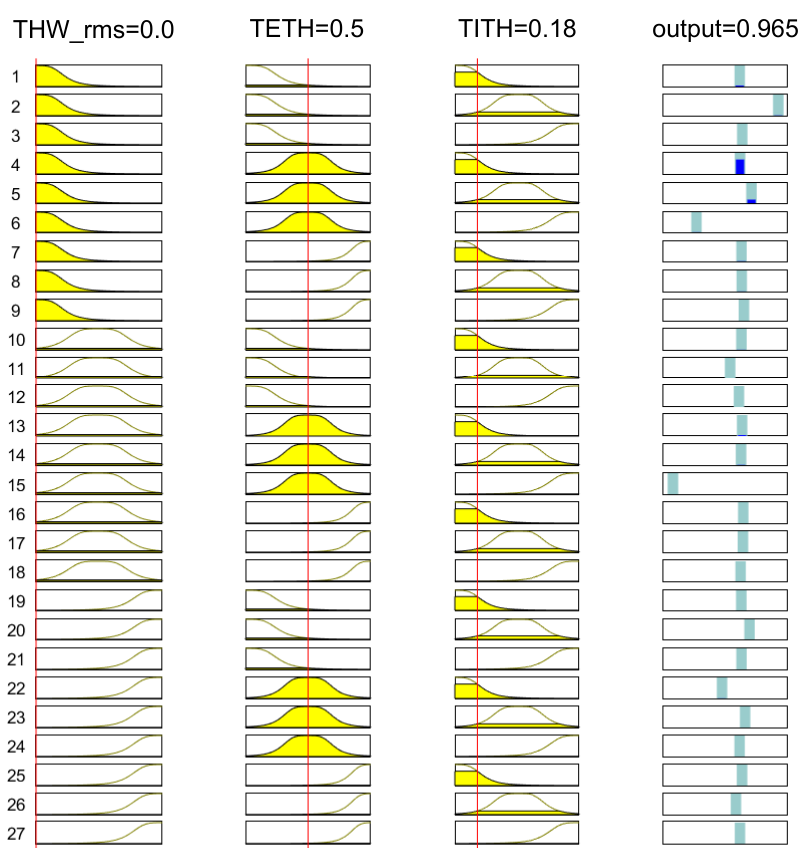}}
 \caption{Rules and membership functions of  ANFIS Cluster 1 for a given input.}
 \label{fig:MATLAB}
 \end{figure}

The particular example of  ANFIS Cluster 1 is displayed in Figure \ref{fig:MATLAB}. In this figure, each column depicts the membership functions for each input of the ANFIS system ($\text{THW}^0_\text{RMS}$, $\text{TETH}^0$, and $\text{TITH}^0$, and  output $y$, respectively), whereas each row corresponds to a fuzzy rule. Thus, for the selected example, with $\text{THW}^0_\text{RMS}=0$, $\text{TETH}^0=0.5$, and $\text{TITH}^0=0.18$, ANFIS Cluster 1 returned an output value $y=0.965$. Since this value was close to 1, and the input data fulfilled the description of Cluster~1 in Section \ref{sec:clustering_results}, this ANFIS correctly identified this value as a member of the cluster it modeled. Additionally, this datapoint was input to the ANFIS of Clusters 2 and 3, with returning output values of 0.017 and 0.31, respectively. As a result, considering the maximum output value of the three ANFIS, the system successfully classified the inputs as Cluster 1.

\begin{figure}[H]
\centering{ \includegraphics[width=1\textwidth, height=0.35\textwidth]{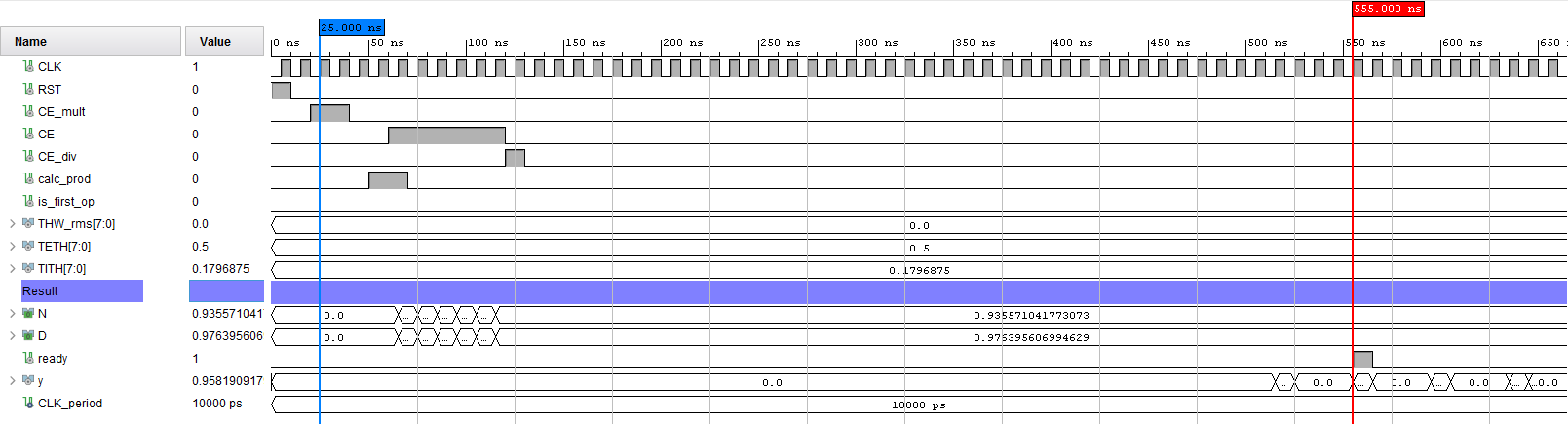}}
 \caption{Simulation results of the ANFIS Cluster 1 HW accelerator obtained with the Vivado design~suite.}
 \label{fig:simu}
 \end{figure}

Regarding  ANFIS HW accelerator verification, in Figure \ref{fig:simu}, a simulation of the ANFIS Cluster 1 HW accelerator is depicted. As can be seen in this figure, with the same input values, the system returned  output $y=0.958$. The results obtained with the HW accelerator agree with Figure \ref{fig:MATLAB}. 

The outputs of the three ANFIS clusters were recovered through the AXI4 bus by the software programmed in the PS partition of the PSoC. The PS determined which  the highest recovered value was, hence identifying the corresponding cluster. 

\subsubsection*{Individual-Based Personalization for ACC ADAS}

Additionally, the software partition in the PS implemented a plane-shaped THW model for each cluster. Thus, given three clusters $1\leq i\leq 3$, the $i$th plane was defined, such that
 \begin{equation}
     \widehat{THW}_i=f\left(\overline{{THW}}_{{RMS}_i},TITH_i\right),
     \label{eq:plane}
 \end{equation}
where $\widehat{THW}_i$ is the individualized THW adjustment, $\overline{THW}_{{RMS}_i}$ is the average $\text{THW}_\text{RMS}$ value observed during the learning period of the system for a particular driver in a steady car-following situation, and $TITH_i$ is the normalized TITH value for the same period.
 
 These planes were defined by the three-point method depending on the minimal, maximal, and average values for $\text{THW}_\text{RMS}$ and TITH for each cluster according to Figure \ref{fig:barras}. With these distributions, the $\overline{THW}_{{RMS}_i}$ and $\sigma_i$ for each cluster were:
 \begin{itemize}[leftmargin=*,labelsep=5.8mm]
    \item {{Cluster }1}: %Please check if italic is necessary.
    $\overline{THW}_{{RMS}_1}=1.08$ s, with $\sigma_1=0.27$ s.
    \item {{Cluster 2}}: $\overline{THW}_{{RMS}_2}=2.44$ s, with $\sigma_2=0.59$ s.
    \item {Cluster 3}: $\overline{THW}_{{RMS}_3}=1.61$ s, with $\sigma_3=0.17$ s.
\end{itemize}

For each cluster, the point of minimum $\text{THW}_\text{RMS}$ and maximum TITH were assigned with a value of $\widehat{THW}_{{RMS}_i}=\overline{THW}_{{RMS}_i}-\sigma_i$, as it corresponded to drivers from that cluster who like to drive with a shorter time gap. On the other hand, drivers who would rather drive with longer time gaps (that is, those who are represented by the point of maximum $\text{THW}_\text{RMS}$ and minimum TITH),  have a value of $\widehat{THW}_{{RMS}_i}=\overline{THW}_{{RMS}_i}+\sigma_i$ assigned.  Finally, intermediate drivers (average values of $\text{THW}_\text{RMS}$ and TITH),  have a value of $\widehat{THW}_{{RMS}_i}=\overline{THW}_{{RMS}_i}$. Consequently,  three points $(\overline{THW}_{{RMS}_i},TITH_i,\widehat{THW}_i)$ that  defined each THW-modeling plane $i$ are as follows.

\begin{itemize}[leftmargin=*,labelsep=5.8mm]
    \item $p_{i1}=(min(THW_{{RMS}_i}), max(TITH_i), \overline{THW}_{{RMS}_i}-\sigma_i)$
    \item $p_{i2}=(max(THW_{{RMS}_i}), min(TITH_i), \overline{THW}_{{RMS}_i}+\sigma_i)$
    \item $p_{i3}=(\overline{THW}_{{RMS}_i}, \overline{TITH}_i, \overline{THW}_{{RMS}_i})$
\end{itemize}

\begin{figure}[H]
    \centering
    \includegraphics[width=0.87\textwidth]{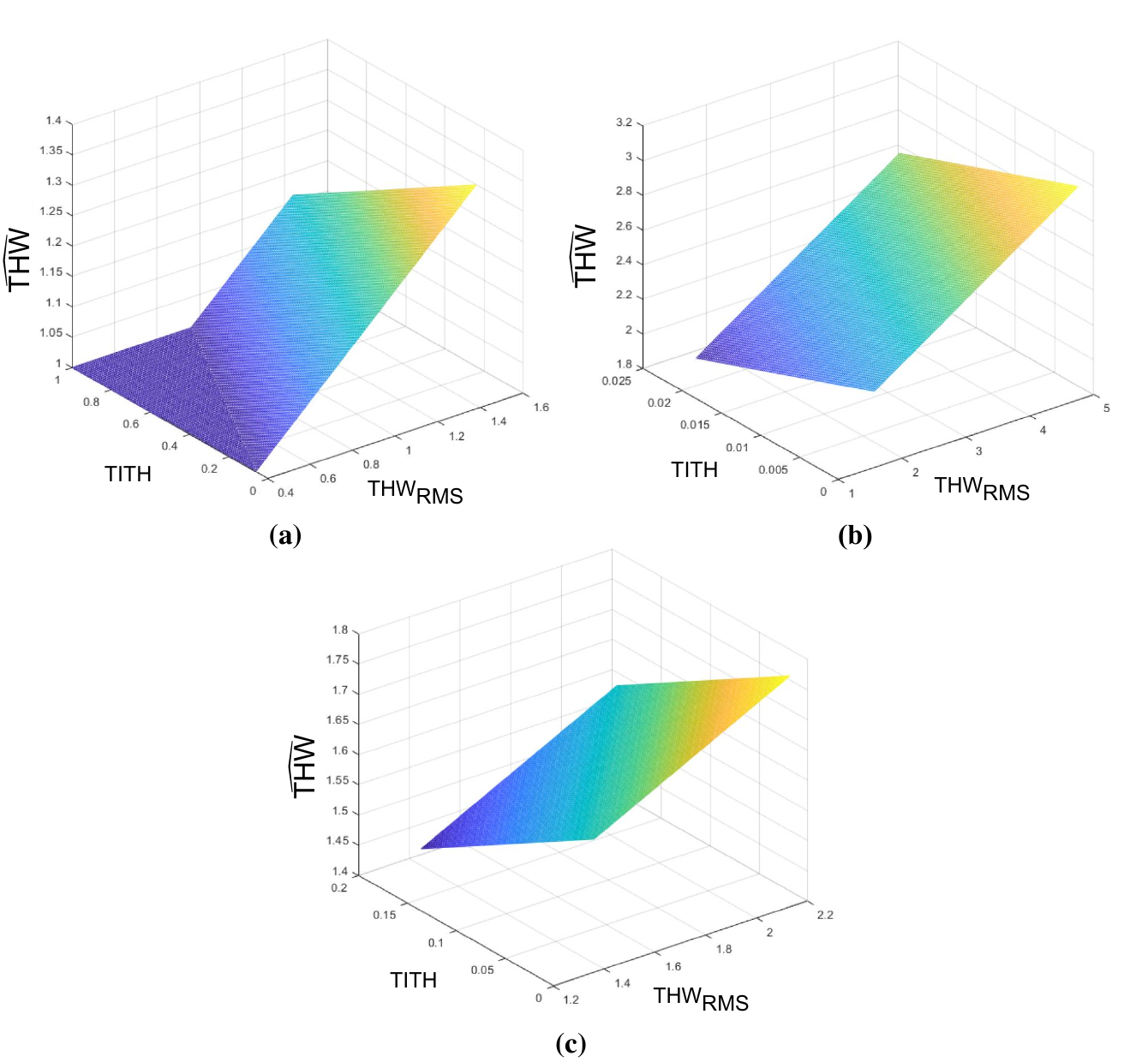}
    \caption{$\widehat{THW}_i$ model planes for  (\textbf{a}) Cluster 1, (\textbf{b}) Cluster 2,  and (\textbf{c})  Cluster 3.}
    \label{fig:planos}
\end{figure}

With these considerations, the planes modeling the individualized $\widehat{THW}_i$ for each cluster according to Equation (\ref{eq:plane}) were computed and shown in Figure \ref{fig:planos}.

As can be seen in Figure \ref{fig:planos}, for each of the models, the predicted $\widehat{THW}_i$ was directly proportional with $\overline{THW}_{{RMS}_i}$ and inversely proportional with TITH. Note that, for the Cluster 1 model, the plane was saturated to $\widehat{THW}_i=1$ s to assure that the personalized THW value never took a value lower than the minimal safe THW values \cite{fleming2018adaptive}.

In sum, once the ANFIS accelerator identified the cluster for a given driver, one of the three models in Figure \ref{fig:planos} was selected. Then, an individualization stage measured and computed the $\text{THW}_\text{RMS}$ and TITH during a steady car-following period. Finally, with those measurements, the system evaluated the corresponding plane model and set a personalized THW value for the ACC system (see Figure \ref{fig:figbloq1}).

\section{Concluding Remarks}
\label{sec:concluding}

In this work, a machine learning approach to face the challenges of ADAS personalization was proposed.  It is based on a hybrid personalization strategy for driving style modeling that uses a group-based clustering technique, namely, k-means clustering  with an individual-based model that adapts the parameters of the clusters to an individual driver. This solution introduces personalization strategies that need no driver intervention with the aim of easing the use of ADAS and thus promoting their adoption in daily driving, with the ultimate goal of increasing road safety and reducing traffic accidents.
The driving style clusters developed in this piece of research are representative of car-following behavior obtained with a meaningful sample of drivers in different kinds of roads, weather conditions, and lighting. Nevertheless, they can easily  be extended to account for the requirements of particular groups of drivers, mainly the most vulnerable drivers (e.g., elderly or inexpert drivers). In addition, a similar approach could be used to personalize and improve current ADAS through different spotlights, such as the fuel economy or passenger comfort.

The implementation of a single-chip driving personalization system for in-car integration requires a high-speed clustering model. The solution adopted in this work relied on high-performance approximation of the clusters using an ANFIS. The universal approximation capability of ANFIS with its inherently parallelizable layered topology make this model suitable for efficient hardware implementation. The whole neuro-fuzzy sensor was  successfully implemented using an FPGA device of a Xilinx Zynq-7000 PSoC providing high speed and low-power consumption for real-time ADAS implementation. In addition, due to the reconfigurable nature of FPGAs, both the hardware  and the software partition of the PSoC could be updated to cope with the continuous changes that new vehicle technologies  introduce.

In future works,  neuro-fuzzy sensor capabilities will be enhanced by broadening the diversity of car-following scenarios. Both acceleration and braking will be analyzed. Moreover, a finer clustering approach will be investigated with the aim of categorizing driving scenarios according to, among others, weather conditions or lighting.

\vspace{6pt} 

%%%%%%%%%%%%%%%%%%%%%%%%%%%%%%%%%%%%%%%%%%
%% optional
%\supplementary{The following are available online at \linksupplementary{s1}, Figure S1: title, Table S1: title, Video S1: title.}


%%%%%%%%%%%%%%%%%%%%%%%%%%%%%%%%%%%%%%%%%%
\authorcontributions{All authors contributed equally to this work}

%%%%%%%%%%%%%%%%%%%%%%%%%%%%%%%%%%%%%%%%%%
\funding{This work was supported in part by the Spanish AEI and European FEDER funds under Grant TEC2016-77618-R (AEI/FEDER, UE)}

%%%%%%%%%%%%%%%%%%%%%%%%%%%%%%%%%%%%%%%%%%
%\acknowledgments{In this section you can acknowledge any support given which is not covered by the author contribution or funding sections. This may include administrative and technical support, or donations in kind (e.g., materials used for experiments).}

%%%%%%%%%%%%%%%%%%%%%%%%%%%%%%%%%%%%%%%%%%
\conflictsofinterest{The authors declare no conflicts of interest.} 

\disclaimer{The findings and conclusions of this paper are those of the authors and do not necessarily represent the views of VTTI, the Transportation Research Board, or the National Academies.}

\reftitle{References}
%\bibliography{biblio}

% The following MDPI journals use author-date citation: Arts, Econometrics, Economies, Genealogy, Humanities, IJFS, JRFM, Laws, Religions, Risks, Social Sciences. For those journals, please follow the formatting guidelines on http://www.mdpi.com/authors/references
% To cite two works by the same author: \citeauthor{ref-journal-1a} (\citeyear{ref-journal-1a}, \citeyear{ref-journal-1b}). This produces: Whittaker (1967, 1975)
% To cite two works by the same author with specific pages: \citeauthor{ref-journal-3a} (\citeyear{ref-journal-3a}, p. 328; \citeyear{ref-journal-3b}, p.475). This produces: Wong (1999, p. 328; 2000, p. 475)

%=====================================
% References, variant B: external bibliography
%=====================================
%\externalbibliography{yes}
%\bibliography{biblio}

%%%%%%%%%%%%%%%%%%%%%%%%%%%%%%%%%%%%%%%%%%
%% optional
%\sampleavailability{Samples of the compounds ...... are available from the authors.}

%% for journal Sci
%\reviewreports{\\
%Reviewer 1 comments and authors’ response\\
%Reviewer 2 comments and authors’ response\\
%Reviewer 3 comments and authors’ response
%}

%%%%%%%%%%%%%%%%%%%%%%%%%%%%%%%%%%%%%%%%%%
\end{document}